\pdfoutput=1

\documentclass[11pt]{article}

\usepackage[preprint]{acl}

\usepackage{times}
\usepackage{latexsym}

\usepackage[T1]{fontenc}

\usepackage[utf8]{inputenc}

\usepackage{microtype}

\usepackage{inconsolata}

\usepackage{kotex} 
\usepackage{amsmath}
\usepackage{amsfonts}
\usepackage{graphicx}
\usepackage{subcaption}
\usepackage{booktabs} 
\usepackage[english]{babel}
\usepackage{amsthm}
\usepackage{thmtools}
\usepackage{multirow}
\usepackage{longtable}
\theoremstyle{plain}
\newtheorem{definition}{Definition}[section]

\usepackage{colortbl}

\title{Repurformer: Transformers for Repurposing-Aware Molecule Generation}

\author{
  Changhun Lee \\
  UNIST \\
  South Korea \\
  \texttt{changhun@unist.ac.kr} \And
  Gyumin Lee \\
  Korea University \\
  South Korea \\
  \texttt{optimizt@korea.ac.kr}
  }

\begin{document}
\maketitle
\begin{abstract}
Generating as diverse molecules as possible with desired properties is crucial for drug discovery research, which invokes many approaches based on deep generative models today. Despite recent advancements in these models, particularly in variational autoencoders (VAEs), generative adversarial networks (GANs), Transformers, and diffusion models, a significant challenge known as \textit{the sample bias problem} remains. This problem occurs when generated molecules targeting the same protein tend to be structurally similar, reducing the diversity of generation. To address this, we propose leveraging multi-hop relationships among proteins and compounds. Our model, Repurformer, integrates bi-directional pretraining with Fast Fourier Transform (FFT) and low-pass filtering (LPF) to capture complex interactions and generate diverse molecules. A series of experiments on BindingDB dataset confirm that Repurformer successfully creates substitutes for anchor compounds that resemble positive compounds, increasing diversity between the anchor and generated compounds.
\end{abstract}

\section{Introduction} \label{sec: intro}
The design of valid and novel molecules with desired biological properties, known as \textit{de novo} molecule generation, is vital to modern drug discovery. Recent advancements in deep generative models, particularly variational autoencoders (VAEs) \cite{kingmaAutoencodingVariationalBayes2022}, generative adversarial networks (GANs) \cite{GoodfellowGAN2014}, Transformers \cite{Vaswani2017attention}, and diffusion models \cite{Ho2020denoising},  have significantly enhanced our ability to generate chemically valid and novel molecules. However, these models need to be further refined to generate molecules that interact with specific target proteins.

Target-specific molecule generation addresses this challenge by producing drug-like molecules that are more likely to bind with specific target proteins \cite{grechishnikovaTransformerNeuralNetwork2021, qianAlphaDrugProteinTarget2022, tan2022}. Nonetheless, there remains a significant issue known as \textit{the sample bias problem}, where reliance on existing protein-compound pairs results in the generation of structurally similar molecules. This phenomenon limits the diversity of generated molecules and hinders the discovery of novel compounds. 

To address this, we propose leveraging multi-hop relationships among proteins and compounds to expand the generative space and increase the diversity of the generated molecules. Our method introduces the concept of repurposing-aware molecule generation, designed to identify and utilize latent multi-hop relations within the protein-compound interaction network.

In this paper, we present Repurformer, a novel model that integrates bi-directional pretraining and advanced signal processing techniques to overcome the limitations of existing models. Repurformer captures complex relationships between proteins and compounds by pretraining encoders in both protein-to-compound and compound-to-protein directions and applying Fast Fourier Transform (FFT) with low-pass filtering (LPF) to the latent space. This approach allows the model to distinguish the different scales of interactions. By focusing on low-frequency components, which correspond to the longer propagation through the multi-hop protein-compound interaction network, Repurformer generates as diverse compounds as possible with desired properties. In summary, the contributions of our work are threefold:
\begin{itemize}
    \item We introduce a framework for repurposing-aware molecule generation to address the \textit{sample bias problem} by leveraging multi-hop relations between proteins and compounds.

    \item We develop Repurformer, a model that integrates a bi-directional pretraining and an FFT-based approach to capture and utilize latent multi-hop relations in an end-to-end manner.

    \item We demonstrate that Repurformer successfully generates valid and diverse molecules, creating substitutes for anchor compounds that resemble positive compounds.
\end{itemize}

\section{Preliminaries} \label{sec: preliminaries}
\subsection{\textit{De novo} Molecule Generation} \label{sec: de-novo-molecule-generation}
\textit{De novo} molecule generation is the process of exploring vast chemical space and producing novel molecules with desired biological properties. With the rapid advancement of artificial intelligence, recent deep generative models have been widely used in molecule generation tasks.

For example, several VAE variants have been introduced thanks to its manipulable latent space, such as charVAE \cite{gomez-bombarelli2018}, SD-VAE \cite{dai2018}, and JT-VAE \cite{jin2018}. GAN has been adopted due to their capability to generate new molecules that are highly similar in structure to existing ones, including ORGAN \cite{guimaraes2018}, ORGANIC \cite{sanchez-lengeling2017}, and Mol-CycleGAN \cite{maziarka2020}. More recently, Transformers and diffusion models have been utilized, based on their success in language modeling and image generation, respectively, such as MolGPT \cite{bagal2022}, MDM \cite{huang2023}, and GeoLDM \cite{xu2023}.

\subsection{Target-specific Molecule Generation} \label{sec: target-specific-molecule-generation}
In drug discovery, identifying drug-target interactions (DTI) is crucial for understanding the bioactivity and therapeutic effects of drugs for specific diseases. Although the deep generative models have proved useful in generating novel and chemically valid molecules, further screening is necessary to evaluate their potential to bind with specific protein targets. Building on this notion, several researchers have developed target-specific molecule generation models to produce novel, drug-like molecules that are highly likely to interact with specific target proteins, including Transformer-based generation \cite{grechishnikovaTransformerNeuralNetwork2021}, AlphaDrug \cite{qianAlphaDrugProteinTarget2022}, SiamFlow \cite{tan2022} and POLYGON \cite{munson2024}.

\subsection{Repurposing-Aware Molecule Generation} \label{sec: repurposing-aware-molecule-generation}
Drug repurposing is a strategy that identifies new therapeutic uses for approved drugs beyond their original indications. This approach offers significant advantages over developing entirely new drug, such as lower failure risk and development costs. The concept of drug repurposing can be defined as multi-hop relationships in the protein-compound interaction network, which is not directly connected but can be accessed through intermediaries. In chemical spaces, proteins and compounds have many-to-many relationships based on their structural coordination. This leads to the assumption that if a compound can reach a specific protein through another compound that shares a common protein (\textit{i.e.,} in the multi-hop relationship), there is potential for repurposing the focal compound.

The repurposability in chemical spaces can introduce a new paradigm for molecule generation, by serving as a key to expanding the generative space and increasing molecular diversity. Previous approaches for target-specific molecule generation tend to generate structurally similar molecules for a specific protein target due to their dependence on known protein-compound interactions. While incorporating randomness in the generation process can contribute to molecular diversity, it may neglect structural coordination with targets, possibly resulting in a trade-off between diversity and binding affinity. In this context, leveraging latent repurposability within the multi-hop relationships among proteins and compounds can provide a reasonable boundary for molecule generation, broadening the generative space and enhancing molecular diversity without sacrificing their drug potency.

\begin{figure*}
    \centering
    \begin{subfigure}[b]{0.7\columnwidth}
        \centering
        \includegraphics[width=\columnwidth]{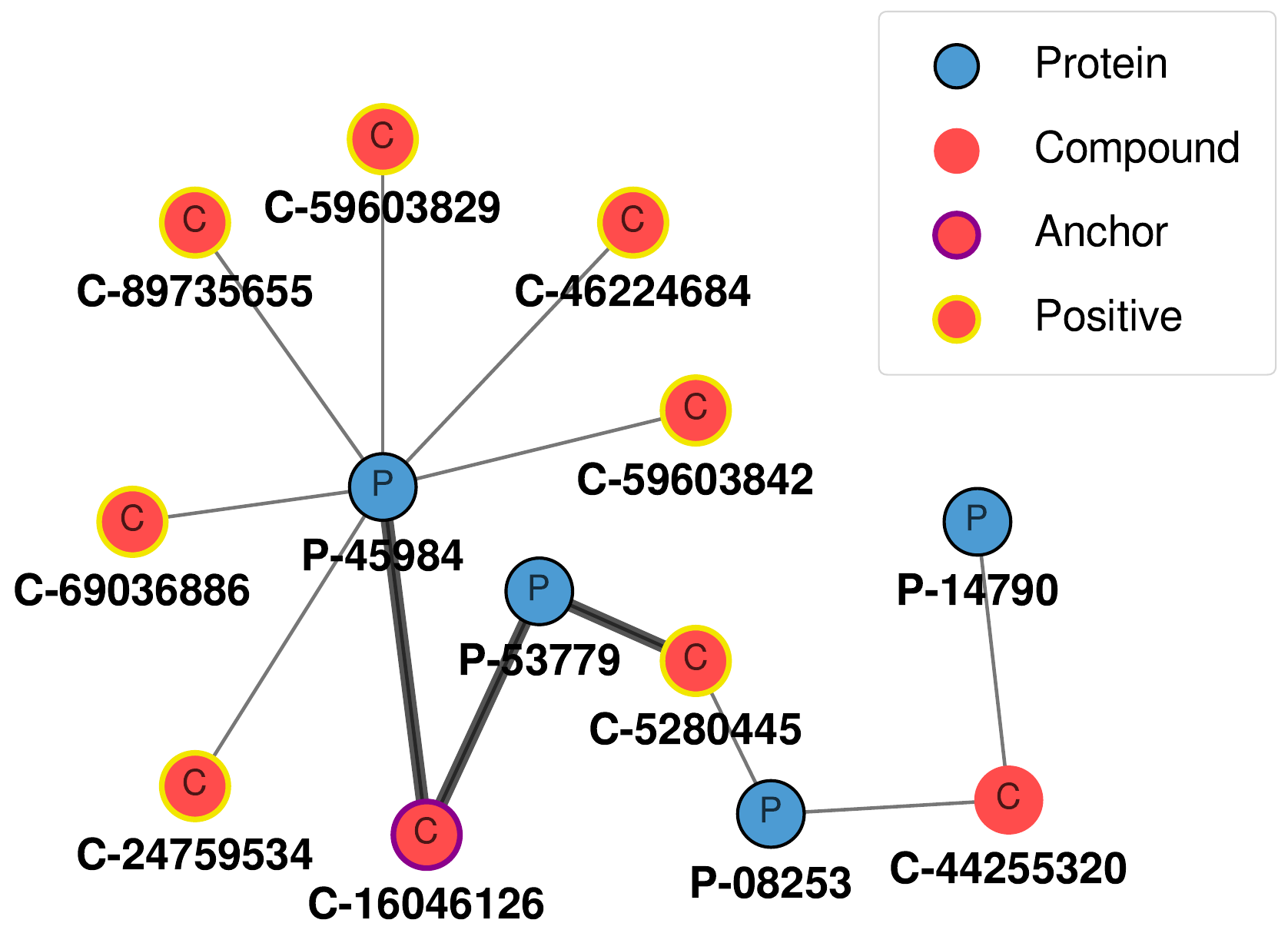}
        \caption{Protein-Compound Graph}
        \label{fig: p-c-graph}
    \end{subfigure}
    \centering
    \hspace{0.025\linewidth}
    \begin{subfigure}[b]{0.6\columnwidth}
        \centering
        \includegraphics[width=\columnwidth]{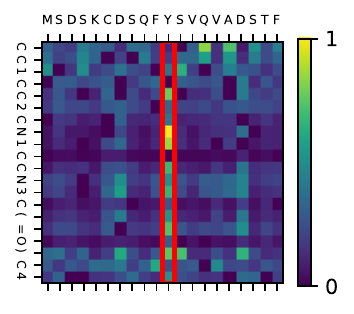}
        \caption{C-16046126 \& P-45984 Attention}
        \label{fig: attention-map-1}

    \end{subfigure}
    \hspace{0.025\linewidth}
    \begin{subfigure}[b]{0.6\columnwidth}
        \centering
        \includegraphics[width=\columnwidth]{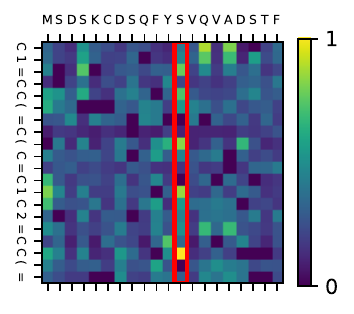}
        \caption{C-5280445 \& P-45984 Attention}
        \label{fig: attention-map-2}
        
    \end{subfigure}
    \caption{(a) illustrates a many-to-many relationship between proteins and compounds. The bold lines indicate potential repurposing flows by which, given an anchor compound's target protein $p$ (P-45984), a positive compound $c^{+}$ (C-5280445) can be considered to replace the anchor compound $\acute{c}$ (C-16046126). Red boxes in (b) and (c) represent the parts of $p$ (P-45984) to which $\acute{c}$ (C-16046126) and $c^{+}$ (C-5280445) attend, respectively. It is noteworthy that attending regions are right next to each other, implying $c^{+}$ may have a potential repurposability to $p$.}
    \label{fig: protein-compound-multi-hop-analysis}
\end{figure*}

\section{Problem Statement} \label{sec: problem-statement}
The discovery of new compounds often relies on existing protein-compound pairs. This results in that the compounds targeting the same protein exhibit similar structures. In other words, the generative space of models tends to be bounded in limited regions, reducing the diversity of the generation. We refer to this as a \textit{sample bias problem}. 

To address this problem, we leverage multi-hop relations among proteins and compounds. Specifically, given a pair of protein $p$ and compound $c$ that are known to interact, we assume that the compound relates to $p$ within a 3-hop relation, \textit{i.e.,} a positive compound $c^{+}$, has a potential interaction with $p$. Definitions from \ref{def: p-c-graph} to \ref{def: anc-pos} describe the key concepts of our approach, and Figure \ref{fig: protein-compound-multi-hop-analysis} visually represents the rationale. Note that both protein and compound are represented by amino-acid and SMILES sequences, respectively: $p = [p_{_{1}}, \cdots, p_{_{T_{p}}}]$ and $c = [c_{_{1}}, \cdots, c_{_{T_{c}}}]$ with $T_{p}$ and $T_{c}$ being the fixed length of each sequence.

\begin{definition}[Protein-Compound Graph] \label{def: p-c-graph}
The relations between proteins and compounds can be represented as a bipartite graph $\mathcal{G}(\mathcal{P} \cup \mathcal{C}, \mathcal{E})$, where $\mathcal{P}$ and $\mathcal{C}$ denote the sets of protein and compound nodes, respectively. Specifically, $p^{(i)} \in \mathcal{P}$ represents the $i$-th protein and $c^{(j)} \in \mathcal{C}$ represents the $j$-th compound, for $i = 1, \cdots, M$ and $j = 1, \cdots, N$.
\end{definition}

\begin{definition}[Protein-Compound Pair] \label{def: p-c-pair}
A pair of nodes in $\mathcal{G}$ is represented by an edge $e_{ij} = \{(p^{(i)}, c^{(j)}) | p^{(i)} \in \mathcal{P}, c^{(j)} \in \mathcal{C}\} \in \mathcal{E}$. The presence of an edge $e_{ij}$ indicates a link between the $i$-th protein and the $j$-th compound, such that $e_{ij} = 1$ if they are linked and $e_{ij} = 0$ otherwise.
\end{definition}

\begin{definition}[Anchor/Positive Compounds] \label{def: anc-pos}
Given a target protein $p^{(i)}$, a compound $c^{(j)}$ is defined as an anchor compound $\acute{c}$ if $e_{ij} = 1$. For another protein $p^{(k)}$ where $e_{kj} = 1$, any compound $c^{(l)}$ ($l \neq j$) that satisfies $e_{kl} = 1$ is regarded as a positive compound $c^{+}$ for the target protein $p^{(i)}$.
\end{definition}

\begin{figure*}[ht]
    \centering
    \begin{subfigure}[b]{0.975\columnwidth}
        \centering
        \includegraphics[width=\linewidth]{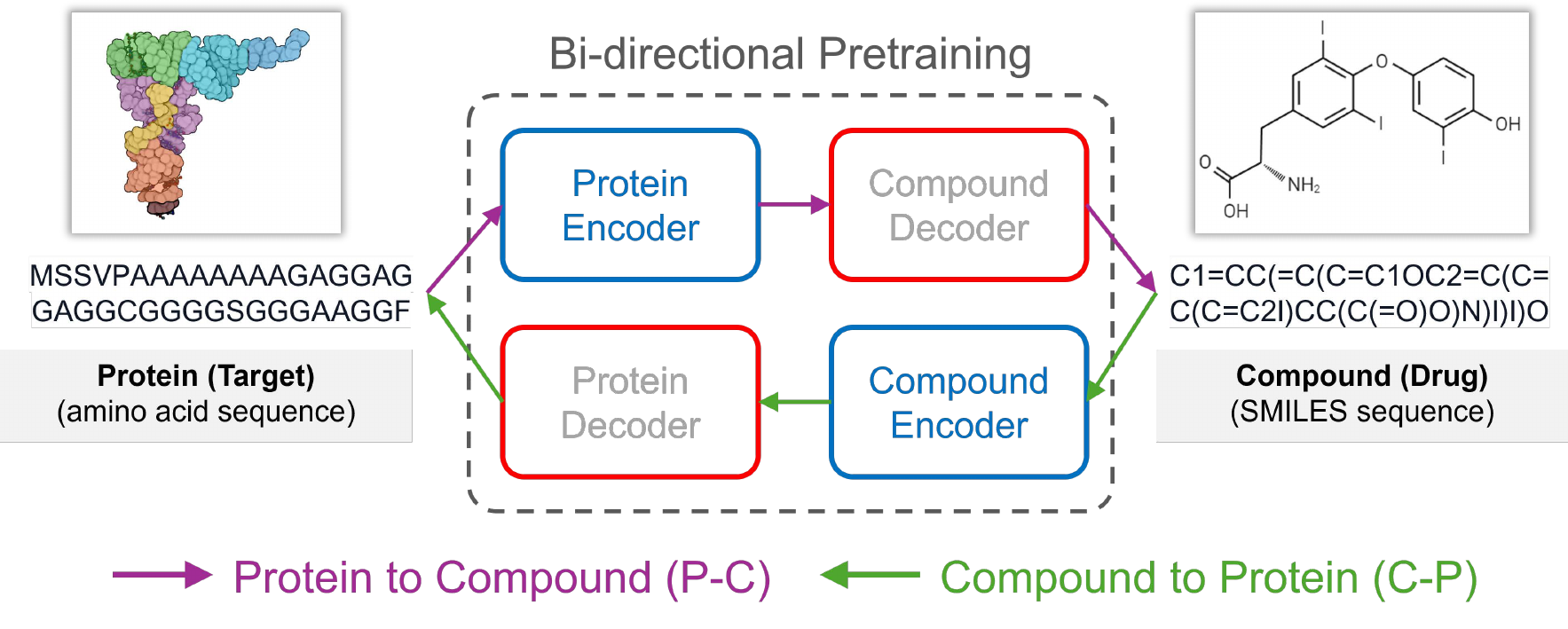}
        \caption{Bi-directioanl Compound-Protein Relations Embedding}
        \label{fig: pretraining-phase}
    \end{subfigure}
    \hspace{0.05\linewidth}
    \begin{subfigure}[b]{0.93\columnwidth}
        \centering
        \includegraphics[width=\linewidth]{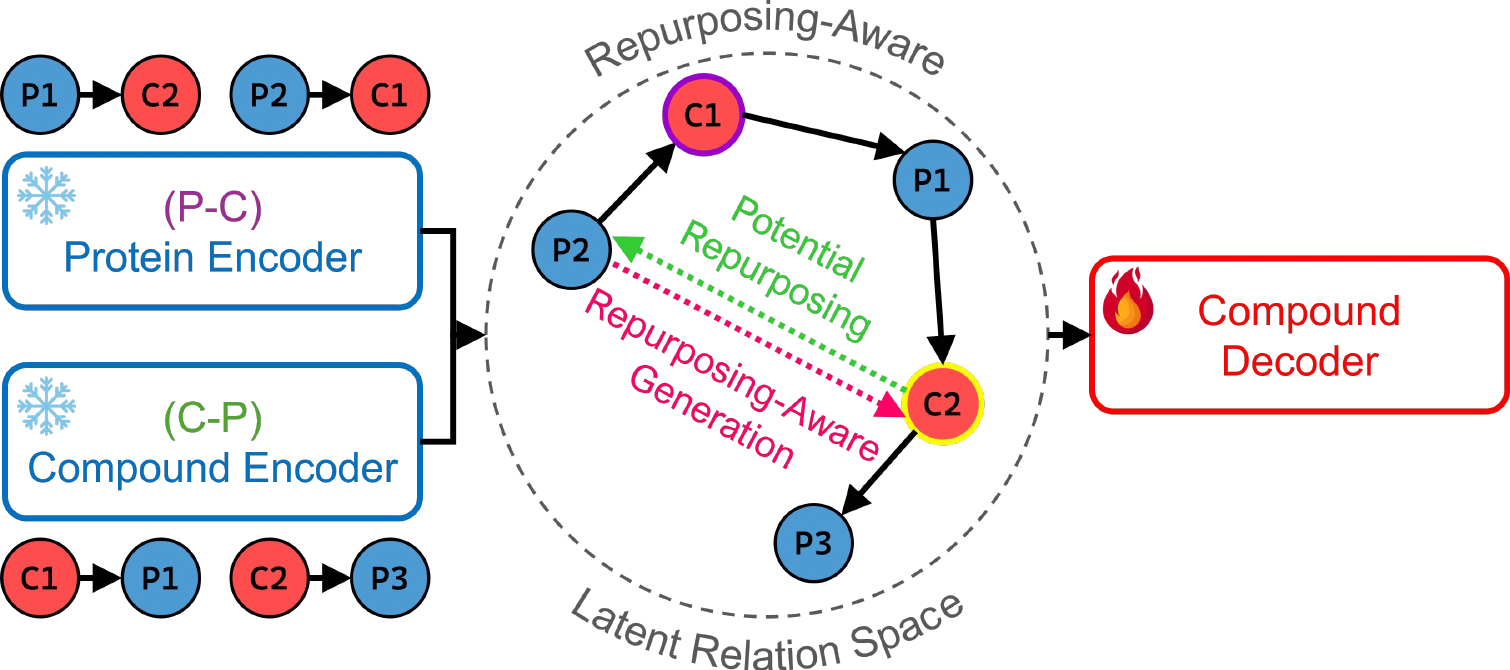}
        \caption{Repurposing-Aware Molecule Generation}
        \label{fig: concept-illustration}
    \end{subfigure}
    \vskip 0.3cm
    \begin{subfigure}[b]{0.95\linewidth}
        \centering
        \includegraphics[width=\linewidth]{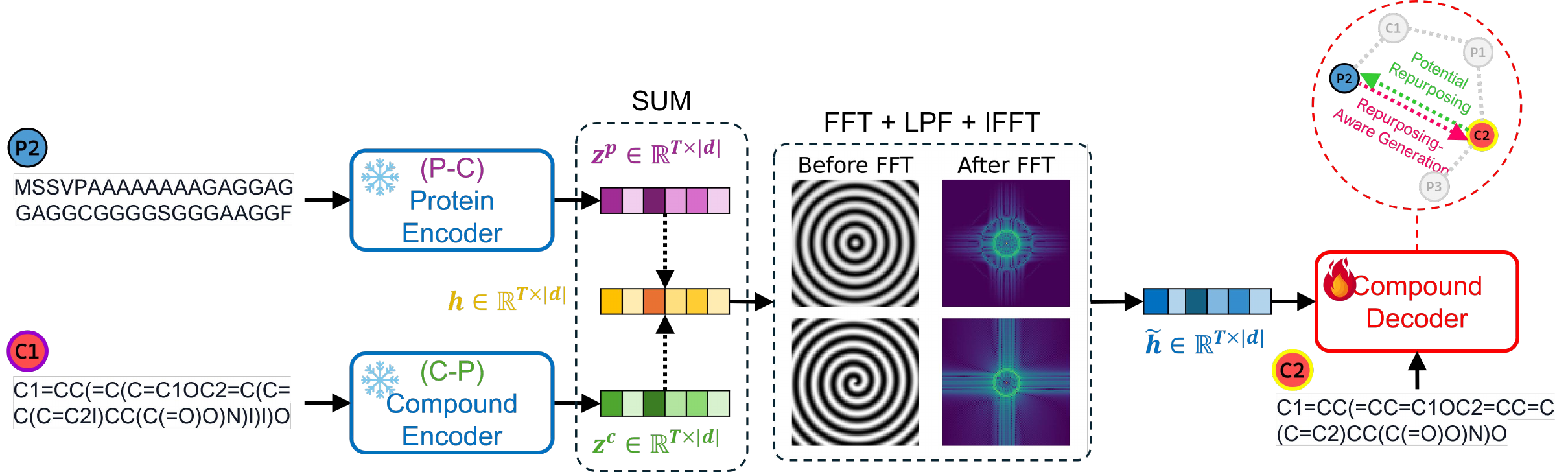}
        \caption{Model Architecture}
        \label{fig: model-architecture}
    \end{subfigure}    
    \caption{Overview of Repurformer}
    \label{fig: overview}
\end{figure*}

\section{Repurformer} \label{sec: repurformer}
In this section, we propose Repurformer, a novel method designed to address the sample bias problem by leveraging multi-hop relations among proteins and compounds. Figure \ref{fig: overview} illustrates how Repurformer seamlessly integrates the concepts of drug discovery and repurposing.

\paragraph{Bi-directional Pretraining}
To capture the many-to-many relationships between proteins and compounds, we employed bi-directional pretraining for the protein and compound encoders. Specifically, we built two Transformers with identical encoder-decoder structures but opposite training directions: one was trained in the protein-to-compound direction, and the other in the compound-to-protein direction (see Figure \ref{fig: pretraining-phase}). By doing so, we expect the protein encoder $f_{p}(c|p)$ and the compound encoder $f_{c}(p|c)$ to extract latent relations, $z^{p}$ and $z^{c}$, that encompass both cases where proteins and compounds are the head and tail of an edge, and vice versa, \textit{i.e.,} $f_{p}: c|p \rightarrow z^{p}$ and $f_{c}: p|c \rightarrow z^{c}$. For example, given a pair of $p^{(2)}$ and $c^{(1)}$ as shown in Figure \ref{fig: concept-illustration}, $z^{p}$ and $z^{c}$ will represent the edges from $p^{(2)}$ to $c^{(1)}$ (\textit{i.e.,} $p^{(2)} \rightarrow c^{(1)}$) and from $c^{(1)}$ to $p^{(1)}$ (\textit{i.e.,} $c^{(1)} \rightarrow p^{(1)}$), respectively.

\paragraph{Transformer with Bi-directional Encoders}
The pretrained bi-encoders are then used as feature extractors; they are frozen and followed by a new compound decoder. The compound decoder $\pi(\cdot)$, parameterized by $\theta$, receives a sum of the encoding vectors $h = z^{p} + z^{c}$ and a positive compound $c^{+}$ as inputs:
\begin{align*}
    \hat{c}^{+}_{t+1} = \pi_{\theta}(\cdot|c^{+}_{1:t}, h_{t}) \quad \text{where} \quad h_{t} = z^{p}_{t} + z^{c}_{t} \ .
\end{align*}
Here, $h_{t} \in \mathbb{R}^{|d|}$ represents a $|d|$-dimensional latent vector of 2-hop relation, $e.g., p^{(2)} \overset{\text{1-hop}}{\rightarrow} c^{(1)} \ (= \acute{c}) \overset{\text{2-hop}}{\rightarrow} p^{(1)}$ (see Figure \ref{fig: concept-illustration}), from a $t$-th token perspective. Accordingly, feeding the compound decoder with a positive compound as a label enables it to learn potential repurposing relationships that emerge from an additional third-hop edge, \textit{e.g.,} $\cdots \overset{\text{2-hop}}{\rightarrow} p^{(1)} \overset{\text{3-hop}}{\rightarrow} c^{(2)} \ (= c^{+})$. Putting it all together, the loss function is defined as follows:
\begin{align*}
    \begin{split}
        \ln \pi_{\theta}(c^{+}|p, c) &= \ln \prod_{t=1}^{T_{c}} \pi_{\theta}(c^{+}_{t+1}|c^{+}_{1:t}, p, \acute{c})        
    \end{split} \\
    \begin{split}
        &= \sum_{t=1}^{T_{c}} \ln \pi_{\theta}(c^{+}_{t+1}|c^{+}_{1:t}, p, \acute{c})
    \end{split}
\end{align*}

\paragraph{Fast Fourier Transform (FFT)}
The Fourier transform decomposes a function into its constituent frequencies using complex exponentials (sinusoids) as basis functions \cite{heckbert1995fourier, lee2021fnet}. Given a sequence $\{ x_{1}, \cdots, x_{T} \}$, the \textit{discrete Fourier Transform} (DFT) is defined by the formula:
\begin{align*}
    X_{k} = \sum_{t=0}^{T-1} x_{t} e^{-\frac{2 \pi i}{T}tk} \ , \quad 0 \leq k \leq T-1
\end{align*}
where \( X_{k} \) is the $k$-th frequency component, \( x_{t} \) is the $t$-th time-domain signal, and $i$ is the imaginary unit. Calculating the DFT directly has a complexity of \(O(T^2)\), which can be inefficient for large datasets. To address this, the \textit{Fast Fourier Transform} (FFT) algorithm was proposed, reducing the complexity to \(O(T \log T)\) \cite{cooley1965algorithm, brigham1988fast}. In this study, we apply the FFT to $h \in \mathbb{R}^{T \times |d|}$ to construct eigenvectors along which the 2-hop propagation occurs. To be specific, the 2D DFT is utilized: one 1D DFT along the sequence dimension, $\mathcal{F}_{\text{seq}}$, and another 1D DFT along the feature dimension, $\mathcal{F}_{\text{dim}}$, keeping real-valued parts only as in \citet{lee2021fnet}:
\begin{align*}
    H = \mathfrak{R}(\mathcal{F}_{\text{seq}}(\mathcal{F}_{\text{dim}}(h))) \in \mathbb{R}^{T \times |d|} \ .
\end{align*}
Note that $T$ is set to the length of a longer sequence; if $T_{p} > T_{c}$, then $T$ is set as $T_{p}$ and vice versa.

\paragraph{Low-Pass Filter (LPF)}
The Fourier-transformed features $H$ comprise low frequencies that represent a globally smoothed signal and high frequencies that indicate a locally normalized signal. This separation of frequency components allows for distinct interpretations at different scales. For example, \citet{tamkin2020language} applied the discrete cosine transform (DCT) \cite{rao2014discrete}, which is closely related to the DFT, to separate latent information at different scales. They found that low frequencies capture topic-scale context while high frequencies capture word-scale context. 

In our setting, a scale can be understood as the number of hops. Specifically, the lower frequency implies a longer propagation through multi-hop relations while the higher one implies a shorter propagation within a single-hop relation. From the repurposing perspective, we need to leverage the longer propagation so that only the multi-hop relations are considered. To achieve this, we can apply the \textit{low-pass filtering} \cite{pollack1948effects, costen1996effects}, which removes the frequency components above a certain cutoff parameter $\alpha$ by setting $H_{k, d} \leftarrow 0$ for all $k, d > \alpha$. This filtering can be easily implemented using a binary mask:
\begin{align*}
    H_{\text{LPF}} = H \odot M
\end{align*}
where $M = \{ m_{t, d} | m_{t, d} \in \{0, 1\}, 1 \leq t \leq T, 1 \leq d \leq |d| \}$ is an one-hot matrix, with $m_{t, d} = 1$ for low-frequency components and $m_{t, d} = 0$ otherwise. Lastly, we transformed $H_{\text{LPF}}$ back to the features of an original domain using the inverse FFT (IFFT), before passing it to the compound decoder:
\begin{align*}
    \tilde{h} = \mathcal{F}_{\text{dim}}^{-1}(\mathcal{F}_{\text{seq}}^{-1}(H_{\text{LPF}})) \in \mathbb{R}^{T \times |d|} \ .
\end{align*}

\paragraph{Implementation Details}
The structure of the Repurformer is essentially identical to that of pretrained transformers. It consists of encoder and decoder networks, each linearly stacked with 4 layers of 256 dimensions, with each layer divided into 4 heads of 64 dimensions. To tokenize the protein and compound sequences, we utilized existing vocabularies from previous works---the protein vocabulary from \citet{rao2019evaluating} and the compound vocabulary from \citet{honda2019smiles}. For training, we set the number of epochs, batch size, and learning rate to 20, 64, and 5e-05, respectively. 


\section{Experiments} \label{sec: experiments}

\paragraph{Experiment Setup}
We collected data from BindingDB \cite{gilson2016} which contains over 2.8 million measured binding affinities of interactions between proteins and drug-like molecules. The collected dataset was then preprocessed to filter out missing values, duplicates, and proteins and compounds with excessively long or short sequences. In particular, given the many-to-many nature of protein-compound relationships, we selected compounds that interact with a reasonable number of individual proteins between 10 and 100, to enable our model to learn various compound structures reacting with different proteins. The resulting dataset comprised 60,719 protein-compound pairs derived from 3,006 proteins and 7,803 compounds. We split this dataset into train and test datasets with 8:2 ratio, ensuring that the proteins interacting with each compound did not overlap between the two sets. Our model was then trained on protein-compound pairs from the train set, representing proteins with amino acid sequences and compounds with canonicalized SMILES strings. We tokenized individual characters from amino acid sequences and SMILES strings, resulting in vocabularies containing 30 characters for proteins and 46 characters for compounds.

\paragraph{Evaluation Metrics}
To thoroughly assess the effectiveness and reliability of Repurformer, we employed several evaluation metrics, focusing on the generative performance of the model and physicochemical properties and drug-likeness of the molecules it generated. In terms of generative performance, we applied widely accepted metrics for sequence generation tasks: BLEU \cite{papineni2002}, GLEU \cite{wuGoogleNeuralMachine2016}, and F1 score of ROUGE \cite{lin2004b}. In particular, we used 1- and 2-gram units as the evaluation basis for these generative metrics. We utilized physicochemical properties, specifically molecular weights and log of octanol-water partition coefficients (LogP) \cite{wildman1999}, to assess the feasibility of molecular structures as drugs. Furthermore, we used other widely used drug-likeness metrics, such as QED \cite{bickerton2012}, SA \cite{ertl2009}, and NP \cite{ertl2008}, to evaluate the potential effectiveness of the generated molecules as drug-like compounds. 

\paragraph{Configurations}
This study aims to analyze whether the configuration of Repurformer is effective. Given that the distinguishing configuration of Repurformer is the application of FFT with LPF in the embedding space, we conducted comparative experiments with different configuration options:
\begin{itemize}
    \item SUM Only: This is the baseline configuration. It directly passes $h$ to the compound decoder. 
    \item +FFT: This configuration transforms $h$ to $H$ but does not revert it to $\tilde{h}$.
    \item +MLP: This configuration adds a single fully-connected layer that mixes the values of $h$ feature-wise.
    \item +FFT+MLP: This configuration mixes the frequencies of $H$.
    \item +FFT+MLP+IFFT w/ auxiliary losses: This configuration mixes the frequencies of $H$ and reverts the mixed $H$ to $\tilde{h}$. Note that L1, L2, and Frobenius norm are added as auxiliary losses to minimize the distance between the MLP output and $\tilde{h}$.
\end{itemize}

\section{Results} \label{sec: results}

\begin{table*}[ht]
\resizebox{\textwidth}{!}{
\begin{tabular}{lc|cccccc|cccccc}
\hline
\multicolumn{2}{c|}{}                                                                            & \multicolumn{6}{c|}{1-gram}                                                                                                                                                                                                                                                                                                                                                                                               & \multicolumn{6}{c}{2-gram}                                                                                                                                                                                                                                                                                                                                                                                                \\ \cline{3-14} 
\multicolumn{2}{c|}{}                                                                            & \multicolumn{2}{c|}{BLEU}                                                                                                                             & \multicolumn{2}{c|}{GLEU}                                                                                                                             & \multicolumn{2}{c|}{ROUGE}                                                                                & \multicolumn{2}{c|}{BLEU}                                                                                                                             & \multicolumn{2}{c|}{GLEU}                                                                                                                             & \multicolumn{2}{c}{ROUGE}                                                                                 \\ \cline{3-14} 
\multicolumn{2}{c|}{\multirow{-3}{*}{}}                                                          & \multicolumn{1}{c|}{anc $\acute{c}$}                                      & \multicolumn{1}{c|}{pos $c^{+}$}                                          & \multicolumn{1}{c|}{anc $\acute{c}$}                                      & \multicolumn{1}{c|}{pos $c^{+}$}                                          & \multicolumn{1}{c|}{anc $\acute{c}$}                                      & pos $c^{+}$                   & \multicolumn{1}{c|}{anc $\acute{c}$}                                      & \multicolumn{1}{c|}{pos $c^{+}$}                                          & \multicolumn{1}{c|}{anc $\acute{c}$}                                      & \multicolumn{1}{c|}{pos $c^{+}$}                                          & \multicolumn{1}{c|}{anc $\acute{c}$}                                      & pos $c^{+}$                   \\ \hline
\rowcolor[HTML]{EFEFEF} 
\multicolumn{2}{l|}{\cellcolor[HTML]{EFEFEF}Baseline (SUM Only)}                                 & \multicolumn{1}{c|}{\cellcolor[HTML]{EFEFEF}0.615}                        & \multicolumn{1}{c|}{\cellcolor[HTML]{EFEFEF}0.664}                        & \multicolumn{1}{c|}{\cellcolor[HTML]{EFEFEF}0.618}                        & \multicolumn{1}{c|}{\cellcolor[HTML]{EFEFEF}0.668}                        & \multicolumn{1}{c|}{\cellcolor[HTML]{EFEFEF}0.381}                        & 0.398                         & \multicolumn{1}{c|}{\cellcolor[HTML]{EFEFEF}0.534}                        & \multicolumn{1}{c|}{\cellcolor[HTML]{EFEFEF}0.580}                        & \multicolumn{1}{c|}{\cellcolor[HTML]{EFEFEF}0.543}                        & \multicolumn{1}{c|}{\cellcolor[HTML]{EFEFEF}0.589}                        & \multicolumn{1}{c|}{\cellcolor[HTML]{EFEFEF}0.120}                        & 0.127                         \\
\multicolumn{2}{l|}{+FFT}                                                                        & \multicolumn{1}{c|}{0.155}                                                & \multicolumn{1}{c|}{0.164}                                                & \multicolumn{1}{c|}{0.179}                                                & \multicolumn{1}{c|}{0.188}                                                & \multicolumn{1}{c|}{0.060}                                                & 0.054                         & \multicolumn{1}{c|}{0.098}                                                & \multicolumn{1}{c|}{0.104}                                                & \multicolumn{1}{c|}{0.126}                                                & \multicolumn{1}{c|}{0.132}                                                & \multicolumn{1}{c|}{0.024}                                                & 0.016                         \\
\rowcolor[HTML]{EFEFEF} 
\multicolumn{2}{l|}{\cellcolor[HTML]{EFEFEF}+MLP}                                                & \multicolumn{1}{c|}{\cellcolor[HTML]{EFEFEF}0.646}                        & \multicolumn{1}{c|}{\cellcolor[HTML]{EFEFEF}{\color[HTML]{FE0000} 0.692}} & \multicolumn{1}{c|}{\cellcolor[HTML]{EFEFEF}0.651}                        & \multicolumn{1}{c|}{\cellcolor[HTML]{EFEFEF}0.700}                        & \multicolumn{1}{c|}{\cellcolor[HTML]{EFEFEF}{\color[HTML]{FE0000} 0.399}} & {\color[HTML]{FE0000} 0.422}  & \multicolumn{1}{c|}{\cellcolor[HTML]{EFEFEF}0.564}                        & \multicolumn{1}{c|}{\cellcolor[HTML]{EFEFEF}{\color[HTML]{FE0000} 0.604}} & \multicolumn{1}{c|}{\cellcolor[HTML]{EFEFEF}0.575}                        & \multicolumn{1}{c|}{\cellcolor[HTML]{EFEFEF}{\color[HTML]{FE0000} 0.618}} & \multicolumn{1}{c|}{\cellcolor[HTML]{EFEFEF}0.133}                        & 0.139                         \\
\multicolumn{2}{l|}{+FFT+MLP}                                                                    & \multicolumn{1}{c|}{0.281}                                                & \multicolumn{1}{c|}{0.289}                                                & \multicolumn{1}{c|}{0.306}                                                & \multicolumn{1}{c|}{0.318}                                                & \multicolumn{1}{c|}{0.011}                                                & 0.012                         & \multicolumn{1}{c|}{0.144}                                                & \multicolumn{1}{c|}{0.156}                                                & \multicolumn{1}{c|}{0.198}                                                & \multicolumn{1}{c|}{0.209}                                                & \multicolumn{1}{c|}{0.004}                                                & 0.004                         \\
\rowcolor[HTML]{EFEFEF} 
\multicolumn{2}{l|}{\cellcolor[HTML]{EFEFEF}+FFT+MLP+IFFT (w/ L1 Loss)}                          & \multicolumn{1}{c|}{\cellcolor[HTML]{EFEFEF}0.583}                        & \multicolumn{1}{c|}{\cellcolor[HTML]{EFEFEF}0.636}                        & \multicolumn{1}{c|}{\cellcolor[HTML]{EFEFEF}0.585}                        & \multicolumn{1}{c|}{\cellcolor[HTML]{EFEFEF}0.640}                        & \multicolumn{1}{c|}{\cellcolor[HTML]{EFEFEF}0.366}                        & 0.388                         & \multicolumn{1}{c|}{\cellcolor[HTML]{EFEFEF}0.511}                        & \multicolumn{1}{c|}{\cellcolor[HTML]{EFEFEF}0.556}                        & \multicolumn{1}{c|}{\cellcolor[HTML]{EFEFEF}0.518}                        & \multicolumn{1}{c|}{\cellcolor[HTML]{EFEFEF}0.565}                        & \multicolumn{1}{c|}{\cellcolor[HTML]{EFEFEF}0.113}                        & 0.113                         \\
\multicolumn{2}{l|}{+FFT+MLP+IFFT (w/ L2 Loss)}                                                  & \multicolumn{1}{c|}{0.623}                                                & \multicolumn{1}{c|}{0.672}                                                & \multicolumn{1}{c|}{0.627}                                                & \multicolumn{1}{c|}{0.679}                                                & \multicolumn{1}{c|}{0.382}                                                & 0.398                         & \multicolumn{1}{c|}{0.543}                                                & \multicolumn{1}{c|}{0.588}                                                & \multicolumn{1}{c|}{0.553}                                                & \multicolumn{1}{c|}{0.600}                                                & \multicolumn{1}{c|}{0.118}                                                & 0.123                         \\
\rowcolor[HTML]{EFEFEF} 
\multicolumn{2}{l|}{\cellcolor[HTML]{EFEFEF}+FFT+MLP+IFFT (w/ Frobenius Loss)}                   & \multicolumn{1}{c|}{\cellcolor[HTML]{EFEFEF}0.629}                        & \multicolumn{1}{c|}{\cellcolor[HTML]{EFEFEF}0.670}                        & \multicolumn{1}{c|}{\cellcolor[HTML]{EFEFEF}0.635}                        & \multicolumn{1}{c|}{\cellcolor[HTML]{EFEFEF}0.679}                        & \multicolumn{1}{c|}{\cellcolor[HTML]{EFEFEF}0.359}                        & 0.367                         & \multicolumn{1}{c|}{\cellcolor[HTML]{EFEFEF}0.544}                        & \multicolumn{1}{c|}{\cellcolor[HTML]{EFEFEF}0.579}                        & \multicolumn{1}{c|}{\cellcolor[HTML]{EFEFEF}0.556}                        & \multicolumn{1}{c|}{\cellcolor[HTML]{EFEFEF}0.594}                        & \multicolumn{1}{c|}{\cellcolor[HTML]{EFEFEF}0.101}                        & 0.104                         \\ \hline
\multicolumn{1}{l|}{}                                       & $\alpha$=2                         & \multicolumn{1}{c|}{0.620}                                                & \multicolumn{1}{c|}{0.660}                                                & \multicolumn{1}{c|}{0.626}                                                & \multicolumn{1}{c|}{0.670}                                                & \multicolumn{1}{c|}{0.331}                                                & 0.348                         & \multicolumn{1}{c|}{0.512}                                                & \multicolumn{1}{c|}{0.548}                                                & \multicolumn{1}{c|}{0.528}                                                & \multicolumn{1}{c|}{0.567}                                                & \multicolumn{1}{c|}{0.102}                                                & 0.112                         \\
\multicolumn{1}{l|}{}                                       & \cellcolor[HTML]{EFEFEF}$\alpha$=4 & \multicolumn{1}{c|}{\cellcolor[HTML]{EFEFEF}{\color[HTML]{FE0000} 0.662}} & \multicolumn{1}{c|}{\cellcolor[HTML]{EFEFEF}0.690}                        & \multicolumn{1}{c|}{\cellcolor[HTML]{EFEFEF}{\color[HTML]{FE0000} 0.670}} & \multicolumn{1}{c|}{\cellcolor[HTML]{EFEFEF}{\color[HTML]{FE0000} 0.703}} & \multicolumn{1}{c|}{\cellcolor[HTML]{EFEFEF}0.385}                        & \cellcolor[HTML]{EFEFEF}0.400 & \multicolumn{1}{c|}{\cellcolor[HTML]{EFEFEF}{\color[HTML]{FE0000} 0.571}} & \multicolumn{1}{c|}{\cellcolor[HTML]{EFEFEF}0.598}                        & \multicolumn{1}{c|}{\cellcolor[HTML]{EFEFEF}{\color[HTML]{FE0000} 0.585}} & \multicolumn{1}{c|}{\cellcolor[HTML]{EFEFEF}0.616}                        & \multicolumn{1}{c|}{\cellcolor[HTML]{EFEFEF}{\color[HTML]{FE0000} 0.147}} & \cellcolor[HTML]{EFEFEF}0.149 \\
\multicolumn{1}{l|}{}                                       & $\alpha$=6                         & \multicolumn{1}{c|}{0.583}                                                & \multicolumn{1}{c|}{0.630}                                                & \multicolumn{1}{c|}{0.587}                                                & \multicolumn{1}{c|}{0.635}                                                & \multicolumn{1}{c|}{0.386}                                                & 0.406                         & \multicolumn{1}{c|}{0.513}                                                & \multicolumn{1}{c|}{0.553}                                                & \multicolumn{1}{c|}{0.521}                                                & \multicolumn{1}{c|}{0.563}                                                & \multicolumn{1}{c|}{0.142}                                                & {\color[HTML]{FE0000} 0.150}  \\
\multicolumn{1}{l|}{\multirow{-4}{*}{+FFT+LPF+IFFT (Ours)}} & \cellcolor[HTML]{EFEFEF}$\alpha$=8 & \multicolumn{1}{c|}{\cellcolor[HTML]{EFEFEF}0.606}                        & \multicolumn{1}{c|}{\cellcolor[HTML]{EFEFEF}0.663}                        & \multicolumn{1}{c|}{\cellcolor[HTML]{EFEFEF}0.610}                        & \multicolumn{1}{c|}{\cellcolor[HTML]{EFEFEF}0.667}                        & \multicolumn{1}{c|}{\cellcolor[HTML]{EFEFEF}0.390}                        & \cellcolor[HTML]{EFEFEF}0.416 & \multicolumn{1}{c|}{\cellcolor[HTML]{EFEFEF}0.532}                        & \multicolumn{1}{c|}{\cellcolor[HTML]{EFEFEF}0.582}                        & \multicolumn{1}{c|}{\cellcolor[HTML]{EFEFEF}0.541}                        & \multicolumn{1}{c|}{\cellcolor[HTML]{EFEFEF}0.591}                        & \multicolumn{1}{c|}{\cellcolor[HTML]{EFEFEF}0.134}                        & \cellcolor[HTML]{EFEFEF}0.144 \\ \hline
\end{tabular}
}
\caption{Evaluation of Generative Performance. The numbers represent the average (n-gram-based) syntactic similarity of the generated compounds $\hat{c}^{+}$, which target specific proteins $p$, to both the anchor compounds $\acute{c}$ and the positive compounds $c^{+}$. Note that $\alpha$ is a cutoff parameter.}
\label{tab: generative-performance}
\end{table*}

\begin{table}[ht]
\resizebox{\columnwidth}{!}{
\begin{tabular}{lc|c|c}
\hline
\multicolumn{2}{c|}{}                                                                              & \begin{tabular}[c]{@{}c@{}}MW\\ $[0, \infty]$\end{tabular} & \begin{tabular}[c]{@{}c@{}}LogP\\ $[-\infty, \infty]$\end{tabular} \\ \hline
\rowcolor[HTML]{EFEFEF} 
\multicolumn{2}{l|}{\cellcolor[HTML]{EFEFEF}Baseline (SUM Only)}                                   & 588.506                                                    & {\color[HTML]{FE0000} 4.870}                                       \\
\multicolumn{2}{l|}{+FFT}                                                                          & \textit{N/A}                                               & \textit{N/A}                                                       \\
\rowcolor[HTML]{EFEFEF} 
\multicolumn{2}{l|}{\cellcolor[HTML]{EFEFEF}+MLP}                                                  & 537.230                                                    & {\color[HTML]{FE0000} 4.317}                                       \\
\multicolumn{2}{l|}{+FFT+MLP}                                                                      & 533.193                                                    & 11.559                                                             \\
\rowcolor[HTML]{EFEFEF} 
\multicolumn{2}{l|}{\cellcolor[HTML]{EFEFEF}+FFT+MLP+IFFT (w/ L1)}                                 & 631.012                                                    & {\color[HTML]{FE0000} 4.655}                                       \\
\multicolumn{2}{l|}{+FFT+MLP+IFFT (w/ L2)}                                                         & 554.490                                                    & {\color[HTML]{FE0000} 4.892}                                       \\
\rowcolor[HTML]{EFEFEF} 
\multicolumn{2}{l|}{\cellcolor[HTML]{EFEFEF}+FFT+MLP+IFFT (w/ Frobenius)}                          & 572.882                                                    & 6.092                                                              \\ \hline
\multicolumn{1}{l|}{}                                         & $\alpha$=2                         & {\color[HTML]{FE0000} 475.473}                             & 6.005                                                              \\
\multicolumn{1}{l|}{}                                         & \cellcolor[HTML]{EFEFEF}$\alpha$=4 & \cellcolor[HTML]{EFEFEF}{\color[HTML]{FE0000} 479.357}     & \cellcolor[HTML]{EFEFEF}{\color[HTML]{FE0000} 3.888}               \\
\multicolumn{1}{l|}{}                                         & $\alpha$=6                         & 584.083                                                    & 7.442                                                              \\
\multicolumn{1}{l|}{\multirow{-4}{*}{+FFT+LPF+IFFT   (Ours)}} & \cellcolor[HTML]{EFEFEF}$\alpha$=8 & \cellcolor[HTML]{EFEFEF}566.942                            & \cellcolor[HTML]{EFEFEF}5.529                                      \\ \hline
\end{tabular}
}
\caption{Evaluation of Physicochemical Properties. The numbers in the MW and LogP columns represent average molecular weights and octanol-water partition coefficients, respectively. By Lipinski's Rule of Five \cite{lipinski2012experimental}, compounds with $\text{MW} \leq 500$ and $\text{LogP} \leq 5$ have good absorption and permeation.}
\label{tab: physicochemical-properties}
\end{table}

\begin{table}[ht]
\resizebox{\columnwidth}{!}{
\begin{tabular}{lc|c|c|c}
\hline
\multicolumn{2}{c|}{}                                                                           & \begin{tabular}[c]{@{}c@{}}QED\\ $[0, 1]$\end{tabular} & \begin{tabular}[c]{@{}c@{}}SA\\ $[1, 10]$\end{tabular} & \begin{tabular}[c]{@{}c@{}}NP\\ $[-5, 5]$\end{tabular} \\ \hline
\rowcolor[HTML]{EFEFEF} 
\multicolumn{2}{l|}{\cellcolor[HTML]{EFEFEF}Baseline (SUM Only)}                                & 0.320                                                  & 4.033                                                  & -0.629                                                 \\
\multicolumn{2}{l|}{+FFT}                                                                       & \textit{N/A}                                           & \textit{N/A}                                           & \textit{N/A}                                           \\
\rowcolor[HTML]{EFEFEF} 
\multicolumn{2}{l|}{\cellcolor[HTML]{EFEFEF}+MLP}                                               & 0.332                                                  & 3.479                                                  & -0.796                                                 \\
\multicolumn{2}{l|}{+FFT+MLP}                                                                   & 0.164                                                  & 2.086                                                  & {\color[HTML]{FE0000} 0.154}                           \\
\rowcolor[HTML]{EFEFEF} 
\multicolumn{2}{l|}{\cellcolor[HTML]{EFEFEF}+FFT+MLP+IFFT (w/ L1)}                              & 0.227                                                  & 4.209                                                  & -0.564                                                 \\
\multicolumn{2}{l|}{+FFT+MLP+IFFT (w/ L2)}                                                      & 0.355                                                  & 3.696                                                  & -0.659                                                 \\
\rowcolor[HTML]{EFEFEF} 
\multicolumn{2}{l|}{\cellcolor[HTML]{EFEFEF}+FFT+MLP+IFFT (w/ Frobenius)}                       & 0.250                                                  & 4.046                                                  & -0.368                                                 \\ \hline
\multicolumn{1}{l|}{}                                      & $\alpha$=2                         & 0.468                                                  & {\color[HTML]{FE0000} 4.289}                           & 0.072                                                  \\
\multicolumn{1}{l|}{}                                      & \cellcolor[HTML]{EFEFEF}$\alpha$=4 & \cellcolor[HTML]{EFEFEF}{\color[HTML]{FE0000} 0.598}   & \cellcolor[HTML]{EFEFEF}2.696                          & \cellcolor[HTML]{EFEFEF}-0.682                         \\
\multicolumn{1}{l|}{}                                      & $\alpha$=6                         & 0.254                                                  & 3.067                                                  & -0.679                                                 \\
\multicolumn{1}{l|}{\multirow{-4}{*}{+FFT+LPF+IFFT(Ours)}} & \cellcolor[HTML]{EFEFEF}$\alpha$=8 & \cellcolor[HTML]{EFEFEF}0.352                          & \cellcolor[HTML]{EFEFEF}3.404                          & \cellcolor[HTML]{EFEFEF}-0.984                         \\ \hline
\end{tabular}
}
\caption{Evaluation of Drug-Likeness. The numbers represent how likely the generated compounds are to be effective drugs. Note that QED, SA, and NP represent a compound's drug-likeness, synthetic accessibility, and natural product-likeness.}
\label{tab: druglikeness}
\end{table}

\begin{table*}[t]
\centering
\resizebox{\textwidth}{!}{
\begin{tabular}{l|cccc|cccc|cc|ccc}
\hline
\multicolumn{1}{c|}{}                   & \multicolumn{4}{c|}{1-gram}                                                                                                                                                                                                                                      & \multicolumn{4}{c|}{2-gram}                                                                                                                                                                                                                                      & \multicolumn{2}{c|}{}                                                                                            & \multicolumn{3}{c}{}                                                                                                                                           \\ \cline{2-9}
\multicolumn{1}{c|}{}                   & \multicolumn{2}{c|}{BLEU}                                                                                                                             & \multicolumn{2}{c|}{GLEU}                                                                                & \multicolumn{2}{c|}{BLEU}                                                                                                                             & \multicolumn{2}{c|}{GLEU}                                                                                & \multicolumn{2}{c|}{\multirow{-2}{*}{\begin{tabular}[c]{@{}c@{}}Physicochemical\\      Properties\end{tabular}}} & \multicolumn{3}{c}{\multirow{-2}{*}{Drug-Likeness}}                                                                                                            \\ \cline{2-14} 
\multicolumn{1}{c|}{\multirow{-3}{*}{}} & \multicolumn{1}{c|}{anc $\acute{c}$}                                      & \multicolumn{1}{c|}{pos $c^{+}$}                                          & \multicolumn{1}{c|}{anc $\acute{c}$}                                      & pos $c^{+}$                  & \multicolumn{1}{c|}{anc $\acute{c}$}                                      & \multicolumn{1}{c|}{pos $c^{+}$}                                          & \multicolumn{1}{c|}{anc $\acute{c}$}                                      & pos $c^{+}$                  & \multicolumn{1}{c|}{MW}                                                        & LogP                            & \multicolumn{1}{c|}{QED}                                                  & \multicolumn{1}{c|}{SA}                            & NP                            \\ \hline
\rowcolor[HTML]{EFEFEF} 
Repurformer ($\alpha$=4)                   & \multicolumn{1}{c|}{\cellcolor[HTML]{EFEFEF}{\color[HTML]{FF0000} 0.662}} & \multicolumn{1}{c|}{\cellcolor[HTML]{EFEFEF}{\color[HTML]{FF0000} 0.690}} & \multicolumn{1}{c|}{\cellcolor[HTML]{EFEFEF}{\color[HTML]{FF0000} 0.670}} & {\color[HTML]{FF0000} 0.703} & \multicolumn{1}{c|}{\cellcolor[HTML]{EFEFEF}{\color[HTML]{FF0000} 0.571}} & \multicolumn{1}{c|}{\cellcolor[HTML]{EFEFEF}{\color[HTML]{FF0000} 0.598}} & \multicolumn{1}{c|}{\cellcolor[HTML]{EFEFEF}{\color[HTML]{FF0000} 0.585}} & {\color[HTML]{FF0000} 0.616} & \multicolumn{1}{c|}{\cellcolor[HTML]{EFEFEF}{\color[HTML]{FF0000} 479.357}}    & {\color[HTML]{FF0000} 3.888}    & \multicolumn{1}{c|}{\cellcolor[HTML]{EFEFEF}{\color[HTML]{FF0000} 0.598}} & \multicolumn{1}{c|}{\cellcolor[HTML]{EFEFEF}2.696} & -0.682                        \\
Transformer                             & \multicolumn{1}{c|}{0.541}                                                & \multicolumn{1}{c|}{0.495}                                                & \multicolumn{1}{c|}{0.599}                                                & 0.572                        & \multicolumn{1}{c|}{0.476}                                                & \multicolumn{1}{c|}{0.440}                                                & \multicolumn{1}{c|}{0.533}                                                & 0.513                        & \multicolumn{1}{c|}{9651.296}                                                  & 187.126                         & \multicolumn{1}{c|}{0.119}                                                & \multicolumn{1}{c|}{{\color[HTML]{FF0000} 9.037}}  & {\color[HTML]{FF0000} -0.128} \\
\rowcolor[HTML]{EFEFEF} 
AlphaDrug                               & \multicolumn{1}{c|}{\cellcolor[HTML]{EFEFEF}0.638}                        & \multicolumn{1}{c|}{\cellcolor[HTML]{EFEFEF}0.652}                        & \multicolumn{1}{c|}{\cellcolor[HTML]{EFEFEF}0.665}                        & 0.685                        & \multicolumn{1}{c|}{\cellcolor[HTML]{EFEFEF}0.555}                        & \multicolumn{1}{c|}{\cellcolor[HTML]{EFEFEF}0.567}                        & \multicolumn{1}{c|}{\cellcolor[HTML]{EFEFEF}{\color[HTML]{FF0000} 0.585}} & 0.603                        & \multicolumn{1}{c|}{\cellcolor[HTML]{EFEFEF}{\color[HTML]{FF0000} 389.616}}    & {\color[HTML]{FF0000} 2.947}    & \multicolumn{1}{c|}{\cellcolor[HTML]{EFEFEF}0.507}                        & \multicolumn{1}{c|}{\cellcolor[HTML]{EFEFEF}2.685} & -0.842                        \\ \hline
\end{tabular}
}
\caption{Evaluation of Comparative Performance. Parts of evaluation metrics in terms of generative performance, physicochemical properties, and drug-likeness are used to compare the performance of Repurformer with the existing target-specific molecule generative models, such as Transformer and AlphaDrug.}
\label{tab: comparative-performance}
\end{table*}

\paragraph{Main Results}
To evaluate Repurformer, we conducted a comparative analysis of 11 configurations, focusing on generative performance, physicochemical properties, and drug-likeness.

Table \ref{tab: generative-performance} shows the similarity of the generated compounds $\hat{c}^{+}$ to both the anchor $\acute{c}$ and positive $c^{+}$ compounds, calculated using BLEU, GLEU, and ROUGE scores. The results indicate that the ``+MLP'' and ``Repurformer with $\alpha=4$'' exhibit remarkable performance compared to other configurations. Notably, the Repurformer ($\alpha=4$) generated compounds with higher structural similarity to the anchor compounds than those generated by the +MLP configuration. This suggests that Repurformer successfully generates compounds that are potentially repurposable to the target proteins.

\begin{figure*}[ht]
    \centering
    \begin{subfigure}[b]{0.485\columnwidth}
        \centering
        \includegraphics[width=\linewidth]{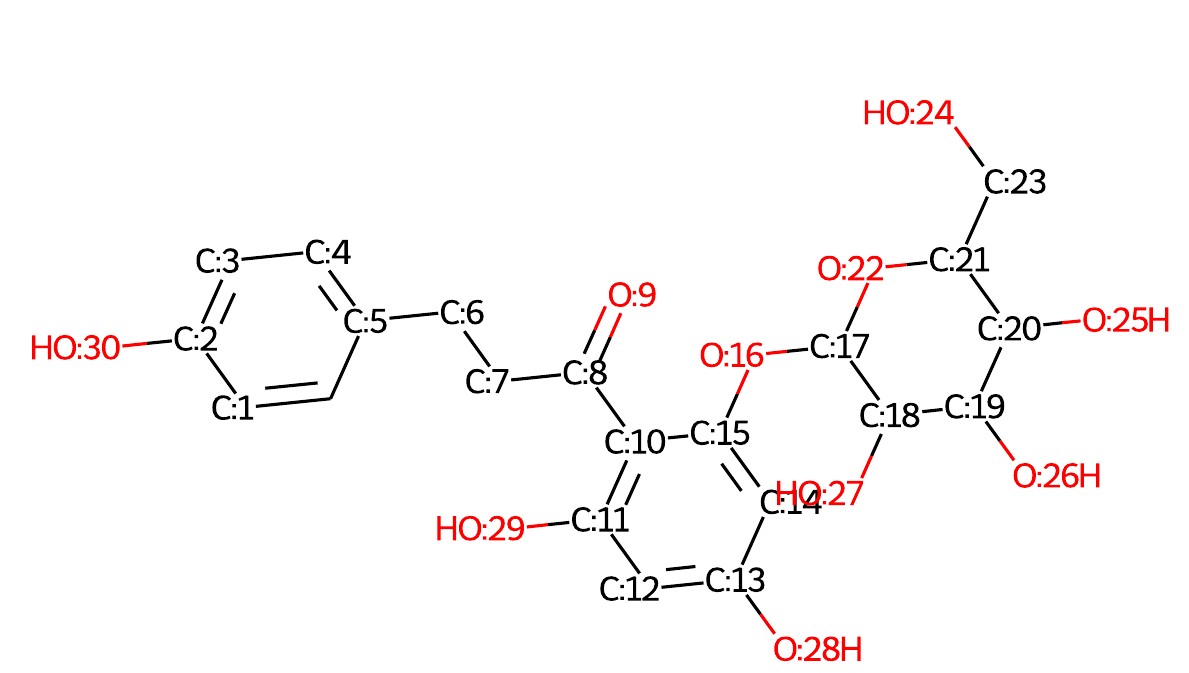}
        \vskip 0.02cm
        \includegraphics[width=\linewidth]{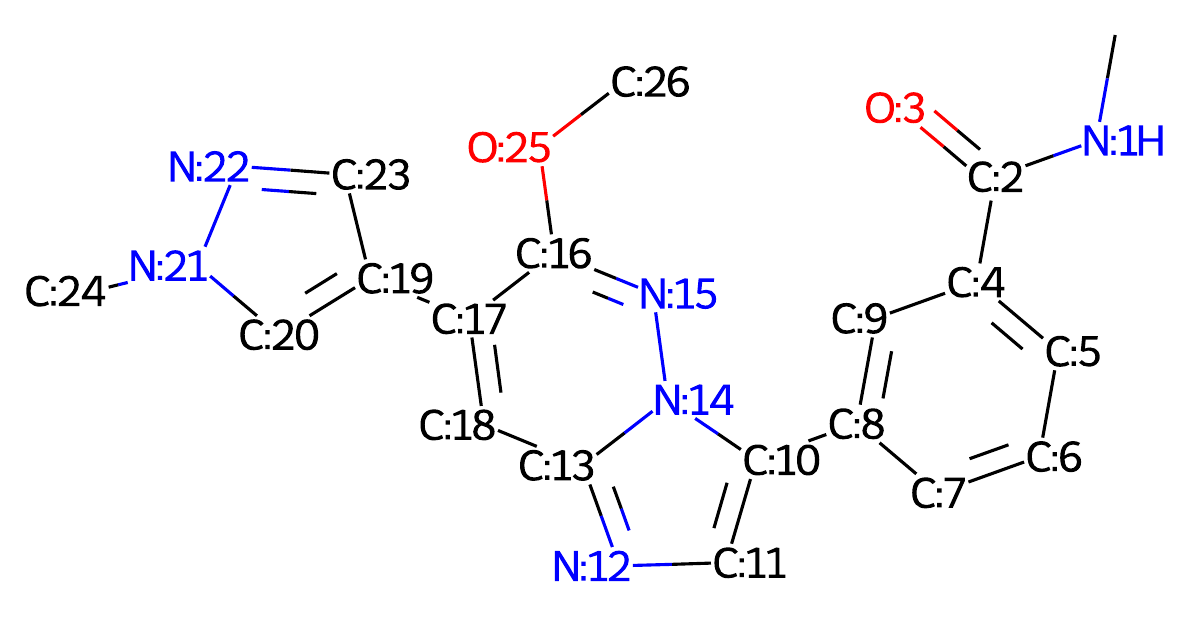}
        \vskip 0.02cm
        \includegraphics[width=\linewidth]{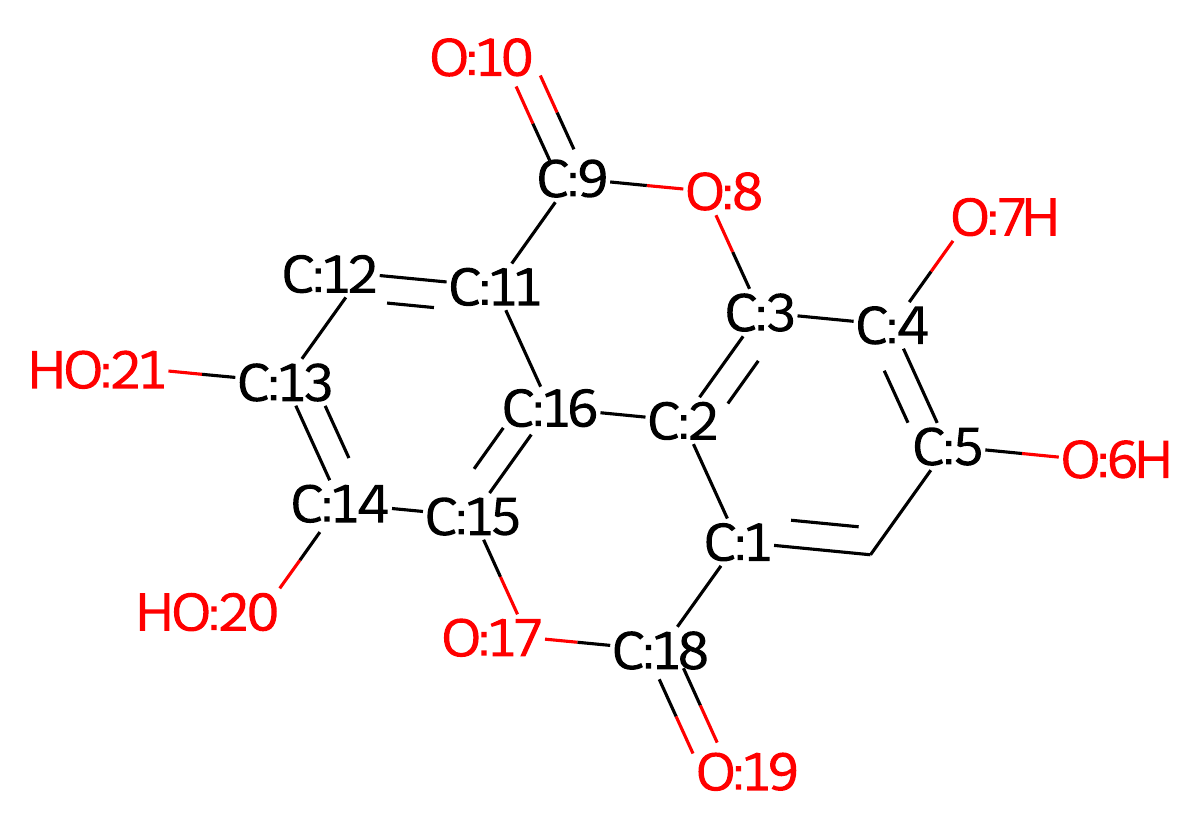}
        \caption{Anchor $\acute{c}$}
    \end{subfigure}
    \hspace{0.01\linewidth}
    \begin{subfigure}[b]{0.485\columnwidth}
        \centering
        \includegraphics[width=\linewidth]{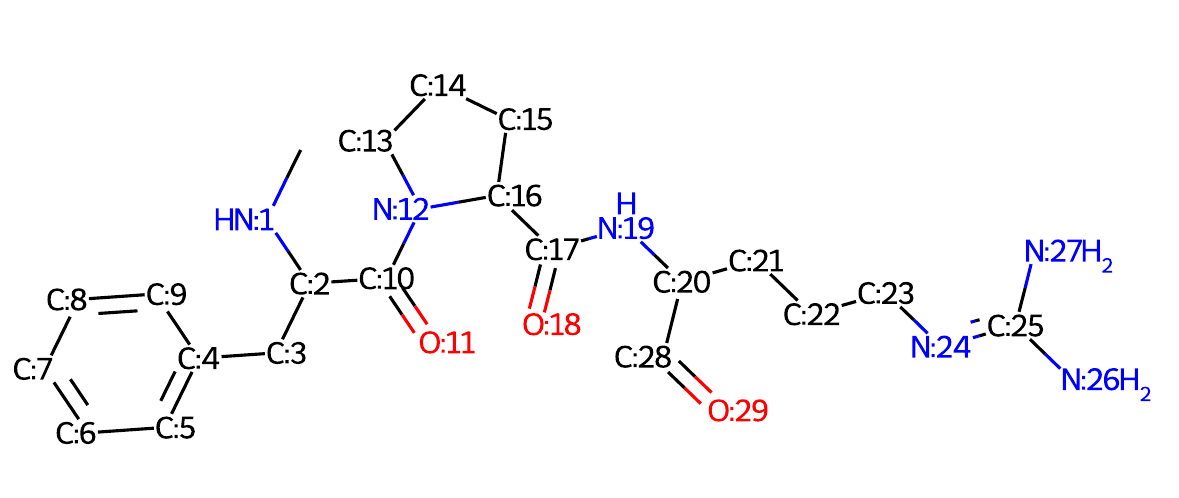}
        \vskip 0.02cm
        \includegraphics[width=\linewidth]{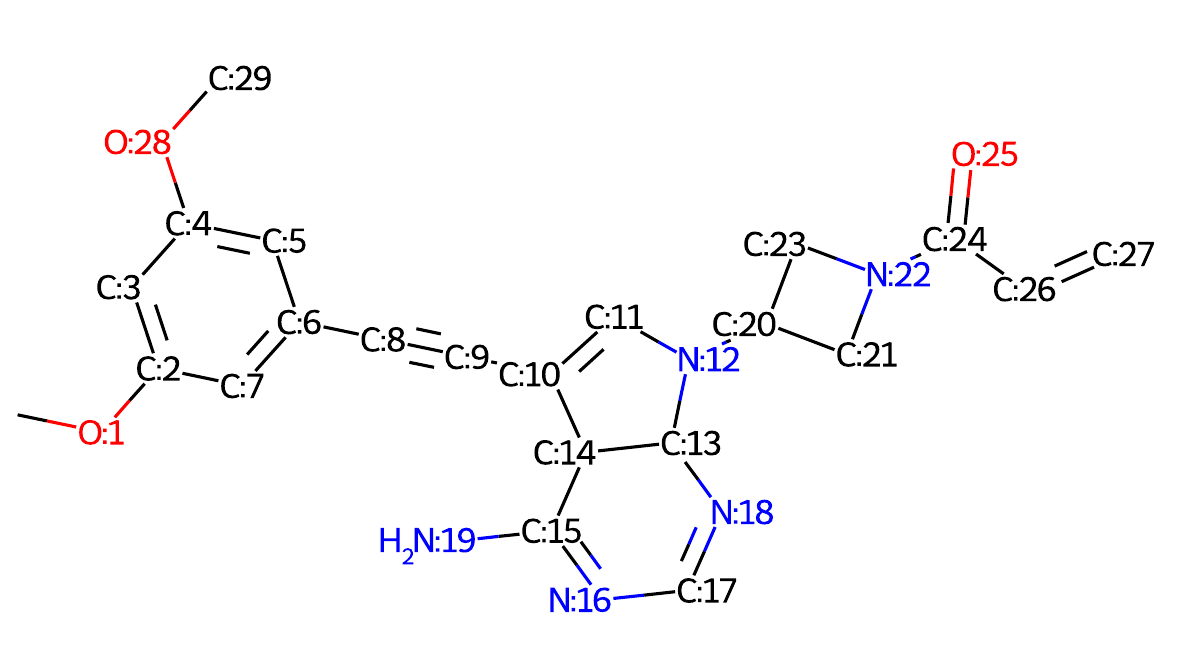}
        \vskip 0.02cm
        \includegraphics[width=\linewidth]{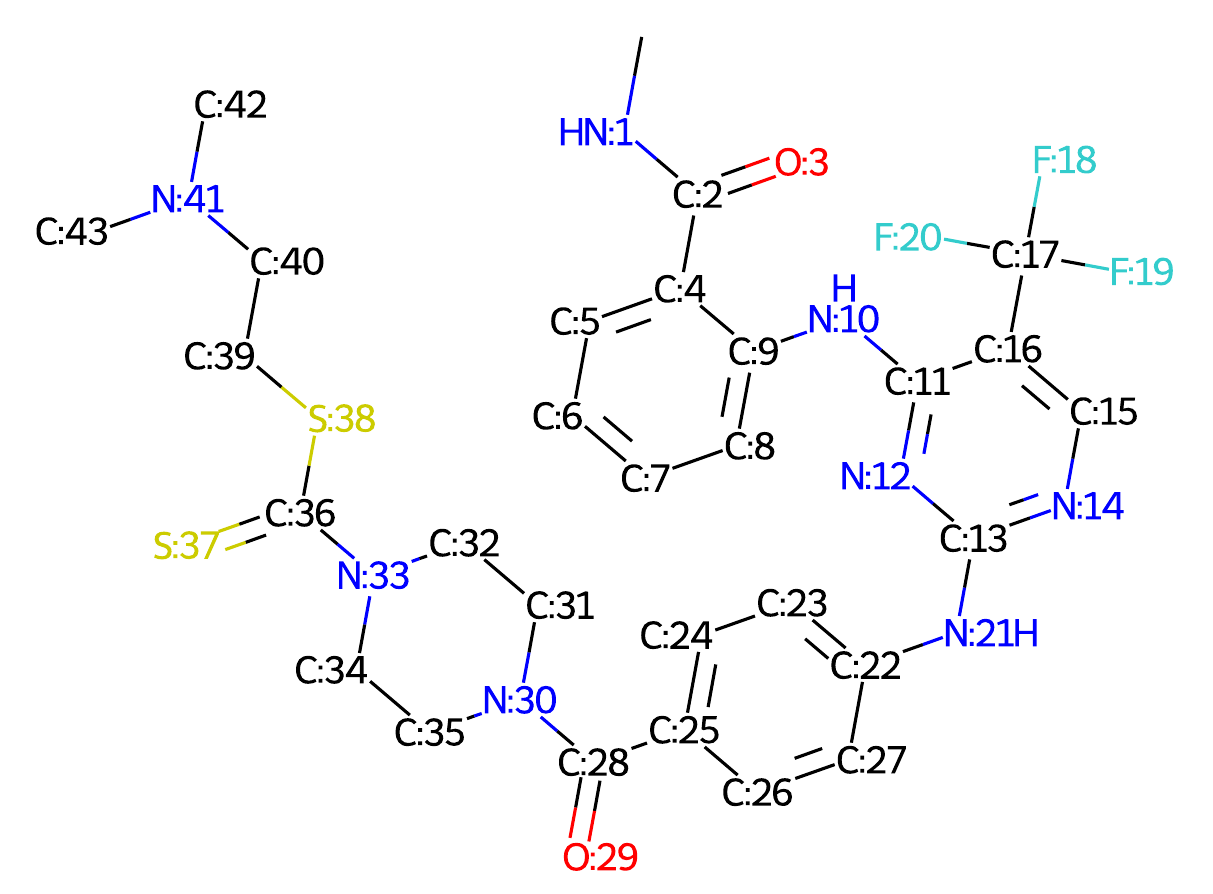}
        \caption{Positive $c^{+}$}
    \end{subfigure}  
    \hspace{0.01\linewidth}
    \begin{subfigure}[b]{0.485\columnwidth}
        \centering
        \includegraphics[width=\linewidth]{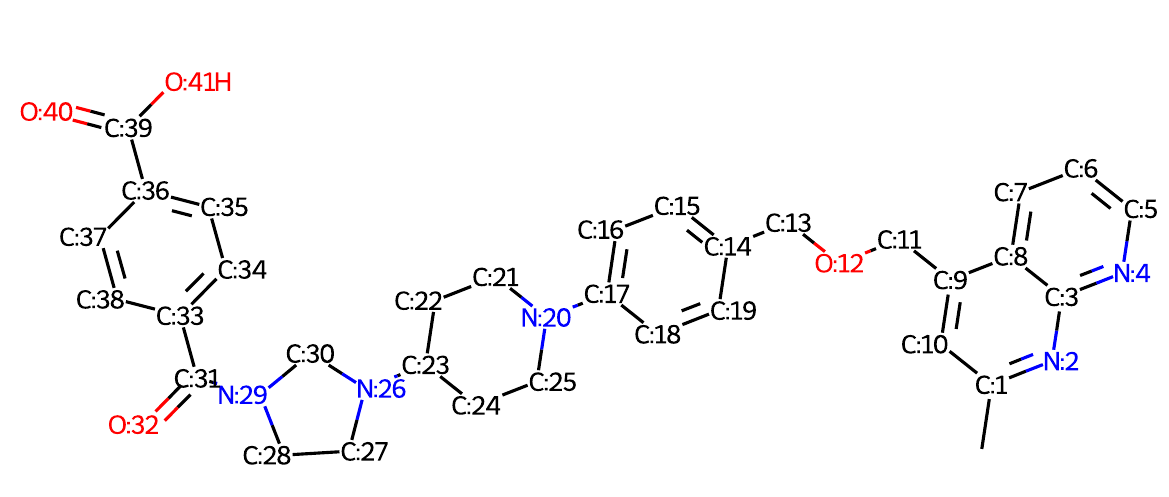}
        \vskip 0.02cm
        \includegraphics[width=\linewidth]{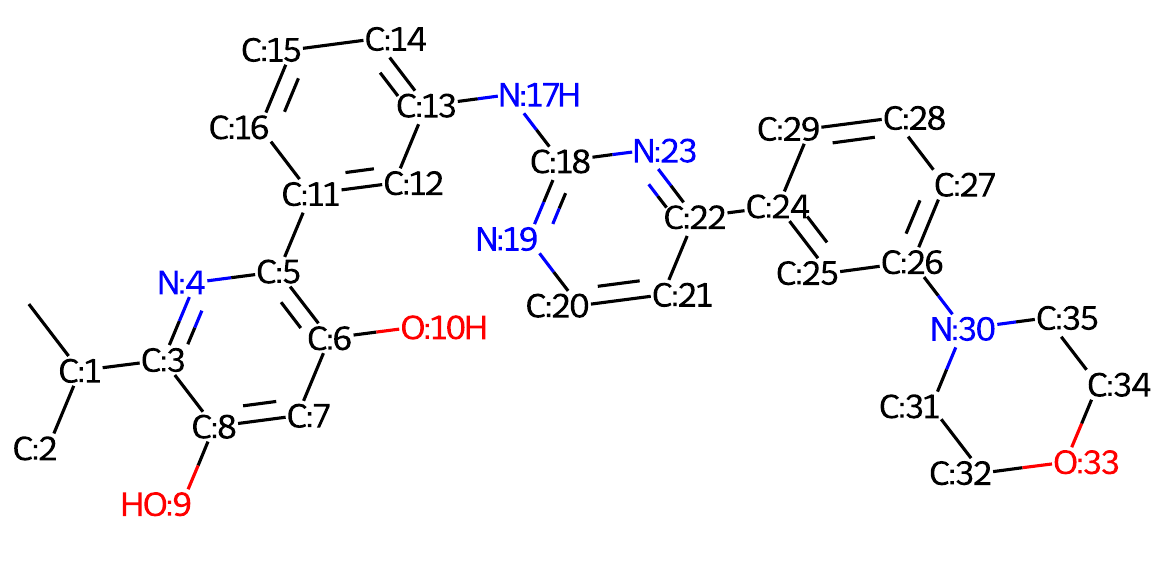}
        \vskip 0.02cm
        \includegraphics[width=\linewidth]{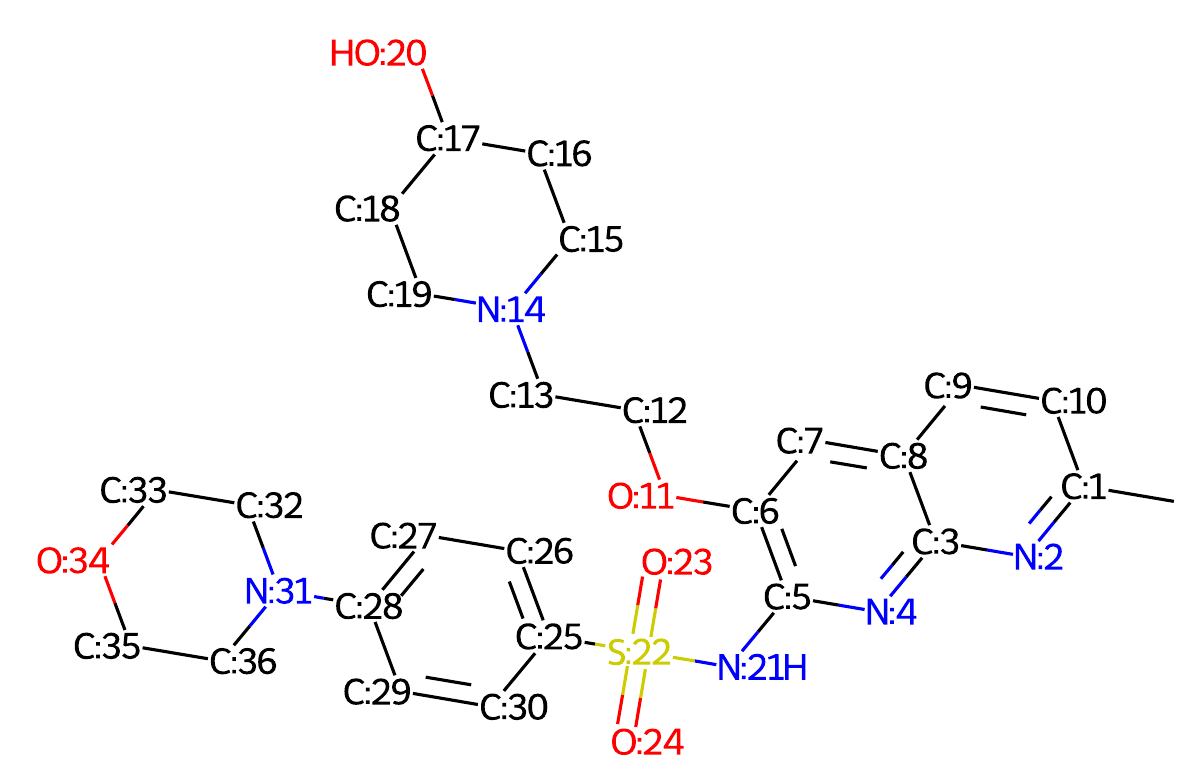}
        \caption{Generated $\hat{c}^{+}$ (+MLP)}
    \end{subfigure}  
    \hspace{0.01\linewidth}
    \begin{subfigure}[b]{0.485\columnwidth}
        \centering
        \includegraphics[width=\linewidth]{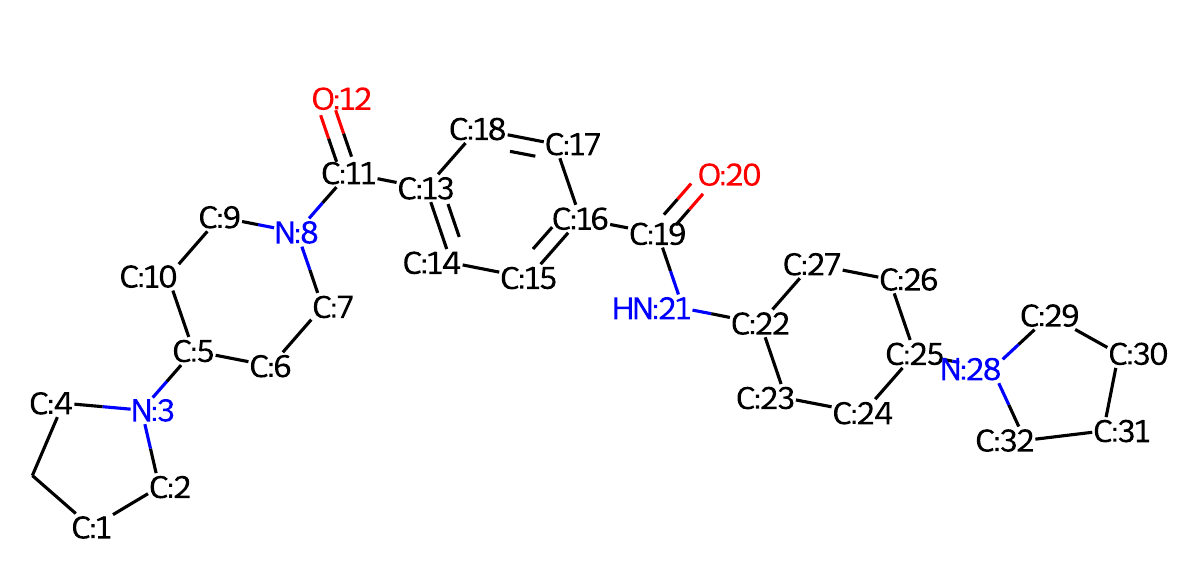}
        \vskip 0.02cm
        \includegraphics[width=\linewidth]{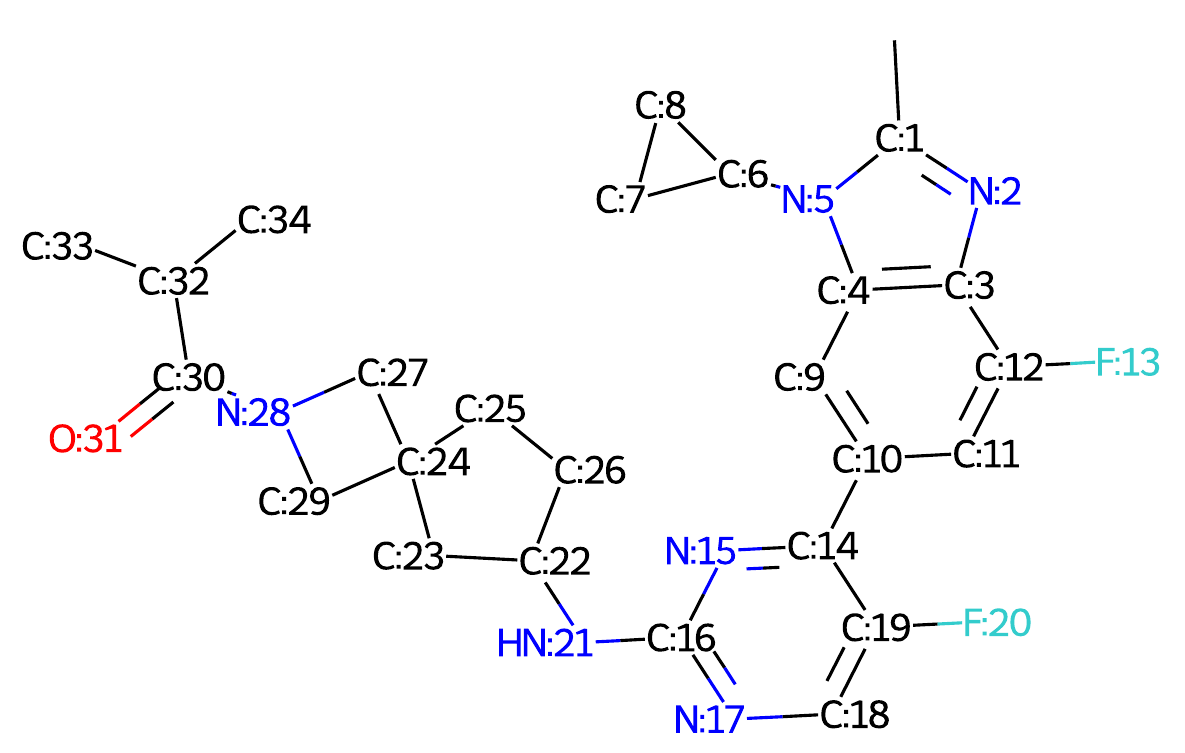}
        \vskip 0.02cm
        \includegraphics[width=\linewidth]{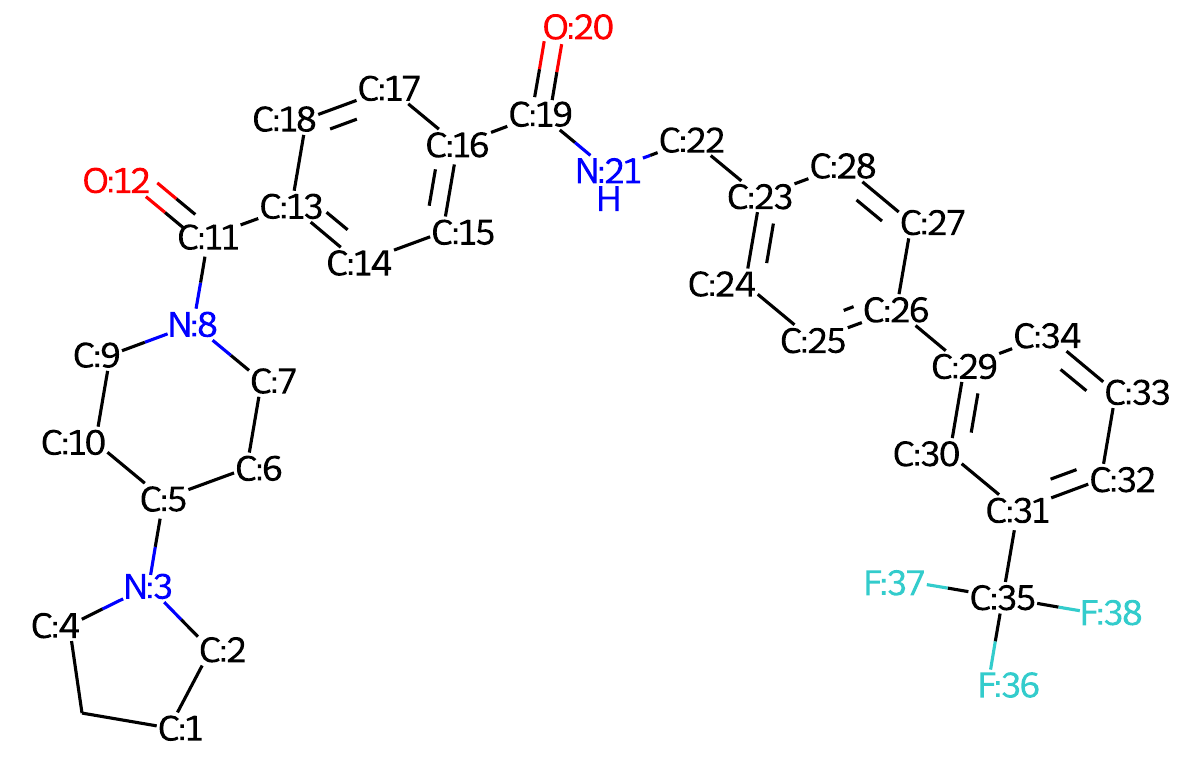}
        \caption{Generated $\hat{c}^{+}$ ($\alpha=4$)}
    \end{subfigure}  
    \caption{Comparison of 2D Molecule Drawings. From left to right, the drawings represent the anchor $\acute{c}$, positive $c^{+}$, and generated compounds $\hat{c}^{+}$, respectively. $\hat{c}^{+}$ is expected to interact with the target protein to which $\acute{c}$ interacts.}
    \label{fig: examples-of-generation}
\end{figure*}

In Tables \ref{tab: physicochemical-properties} and \ref{tab: druglikeness}, we can compare the molecular properties of the generated compounds from different configurations. Table \ref{tab: physicochemical-properties} shows that Repurformer with $\alpha=4$ generates compounds that are the most physicochemically desirable. On the other hand, Table \ref{tab: druglikeness} shows that the Repurformer with $\alpha=4$, with $\alpha=2$, and ``+FFT+MLP'' configurations had comparative advantages in QED, SA, and NP, respectively. Given that QED is generally considered the most important metric for measuring drug similarity and efficacy, we can emphasize that Repurformer ($\alpha=4$) excels in generating compounds with the highest potential for effective drug discovery. Figure \ref{fig: examples-of-generation} compares the generation results of the `+MLP' and `Repurformer ($\alpha=4$)' configurations.

\begin{figure*}[h]
    \begin{subfigure}[b]{0.5\linewidth}
        \centering
        \includegraphics[width=1.0\linewidth]{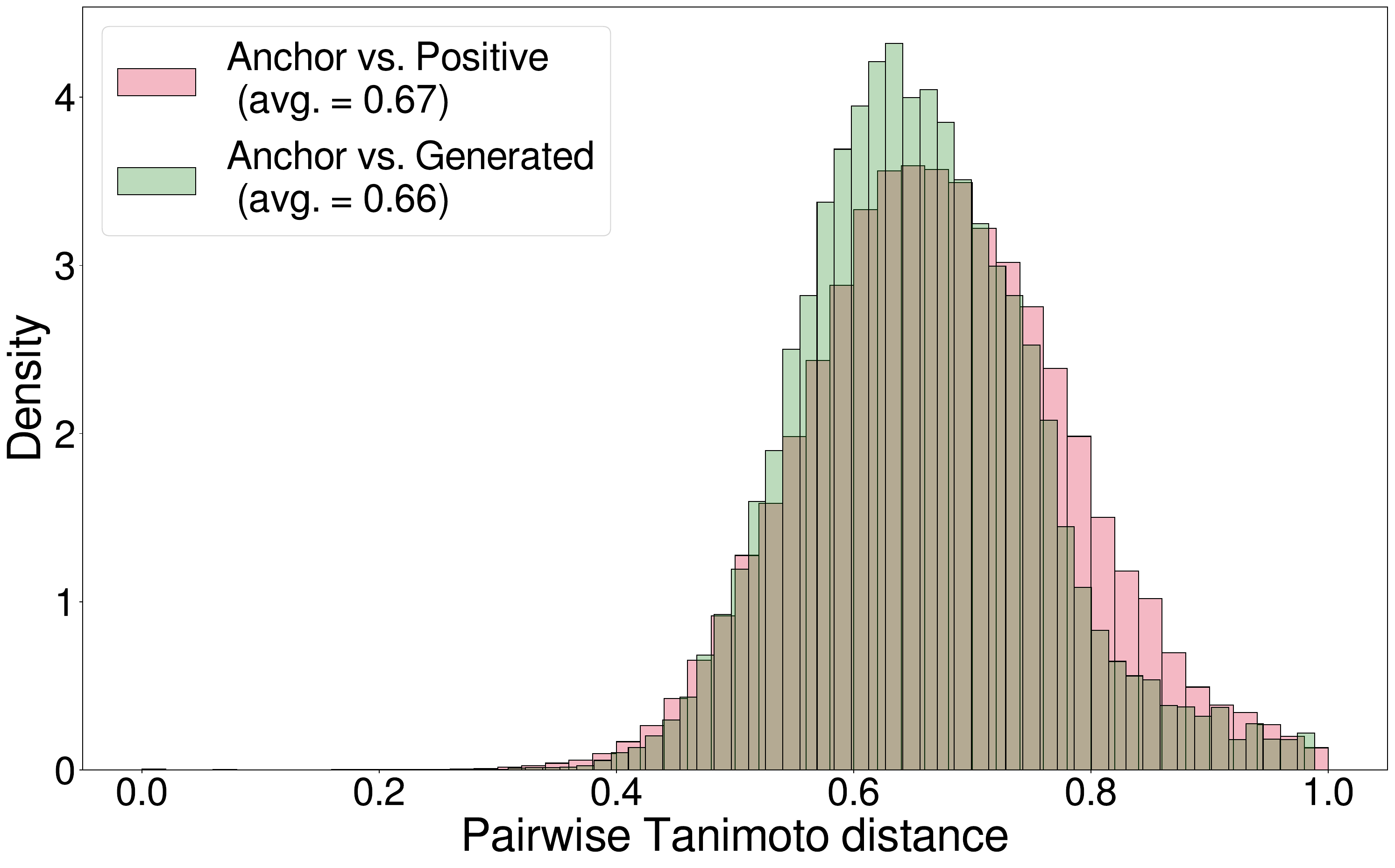}
        \caption{Distribution of Pairwise Tanimoto Distances: Generated vs. Anchor and Positive vs. Anchor. The distance was defined over the molecular fingerprints domain.}
        \label{fig: tanimoto-plot}
    \end{subfigure}
    \hspace{0.01\linewidth}    
    \begin{subfigure}[b]{0.5\linewidth}
        \centering
        \includegraphics[width=0.8\linewidth]{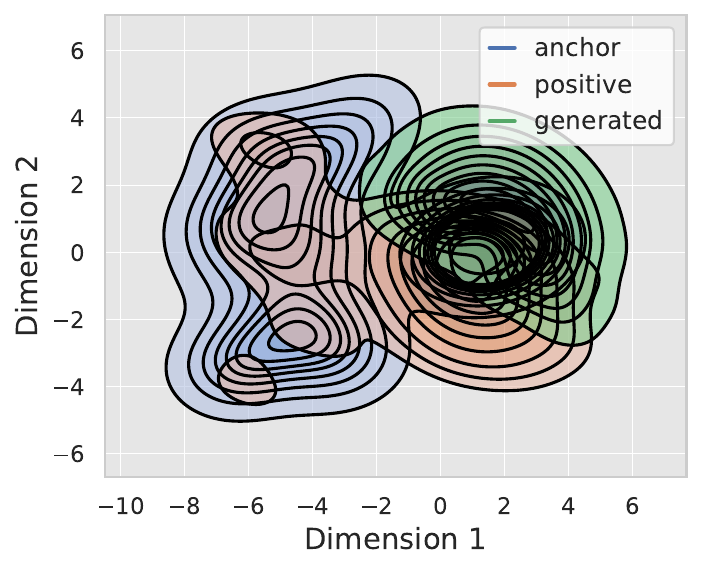}
        \caption{2D Gaussian KDE plots for the anchor, positive, and generated compounds. Before employing KDE, each compound was converted to an embedding vector using successive applications of Word2Vec and t-SNE embeddings.}
        \label{fig: kde-plot}
    \end{subfigure}
    \caption{(a) illustrates the distance distribution from the molecular fingerprint perspective. (b) describes the estimated two-dimensional Gaussian distribution of anchor, positive, and generated compounds.}
\end{figure*}


\paragraph{Performance Comparison}
To assess the effectiveness of Repurformer as a target-specific molecule generation model, we compared its performance with the existing protein-specific generative approaches as external baseline models, including Transformer-based model \cite{grechishnikovaTransformerNeuralNetwork2021} and AlphaDrug \cite{qianAlphaDrugProteinTarget2022}. Transformer-based model utilized the vanilla Transformer architecture \cite{Vaswani2017attention} to generate compounds based on target proteins. This model viewed the target-specific molecule generation as a translational task, converting amino acid sequence into SMILES strings. AlphaDrug modified the vanilla Transformer by introducing skip-connections between its encoders and decoders, facilitating the joint embedding of target proteins and molecules. In addition, it employed a Monte Carlo tree search algorithm for the conditioned generation of novel molecules based on specific target proteins.

To ensure a fair comparison, we trained the external baseline models using our dataset using the same experiment setup and evaluation metrics as for Repurformer. Table 4 presents the performance comparison between our best configuration (Repurformer with $\alpha$ = 4) and the external baseline models. The comparison results demonstrate that the Repurformer ($\alpha$ = 4) outperformed the existing approaches on most evaluation metrics. In particular, our model generated compounds with high structural similarity to both the anchor and positive compounds than those generated by the external baseline models. This suggests that Repurformer can generate not only realistic but diverse compounds with methodological considerations for drug repurposability. Regarding drug-likeness, our model achieved highest performance only on QED. Although the Transformer-based model excelled in SA and NP, the feasibility of its generated compounds is questionable due to its exceptionally high scores in physicochemical properties, which indicate the compounds might not be suitable as medicines. This is further validated by the evaluation of compound validity, as illustrated in Figure \ref{fig: radar-chart} in the Appendix. The compounds generated by the Transformer-based model were significantly less valid compared to those generated by Repurformer.

\begin{figure}[t]  
    \centering
    \includegraphics[width=0.95\columnwidth]{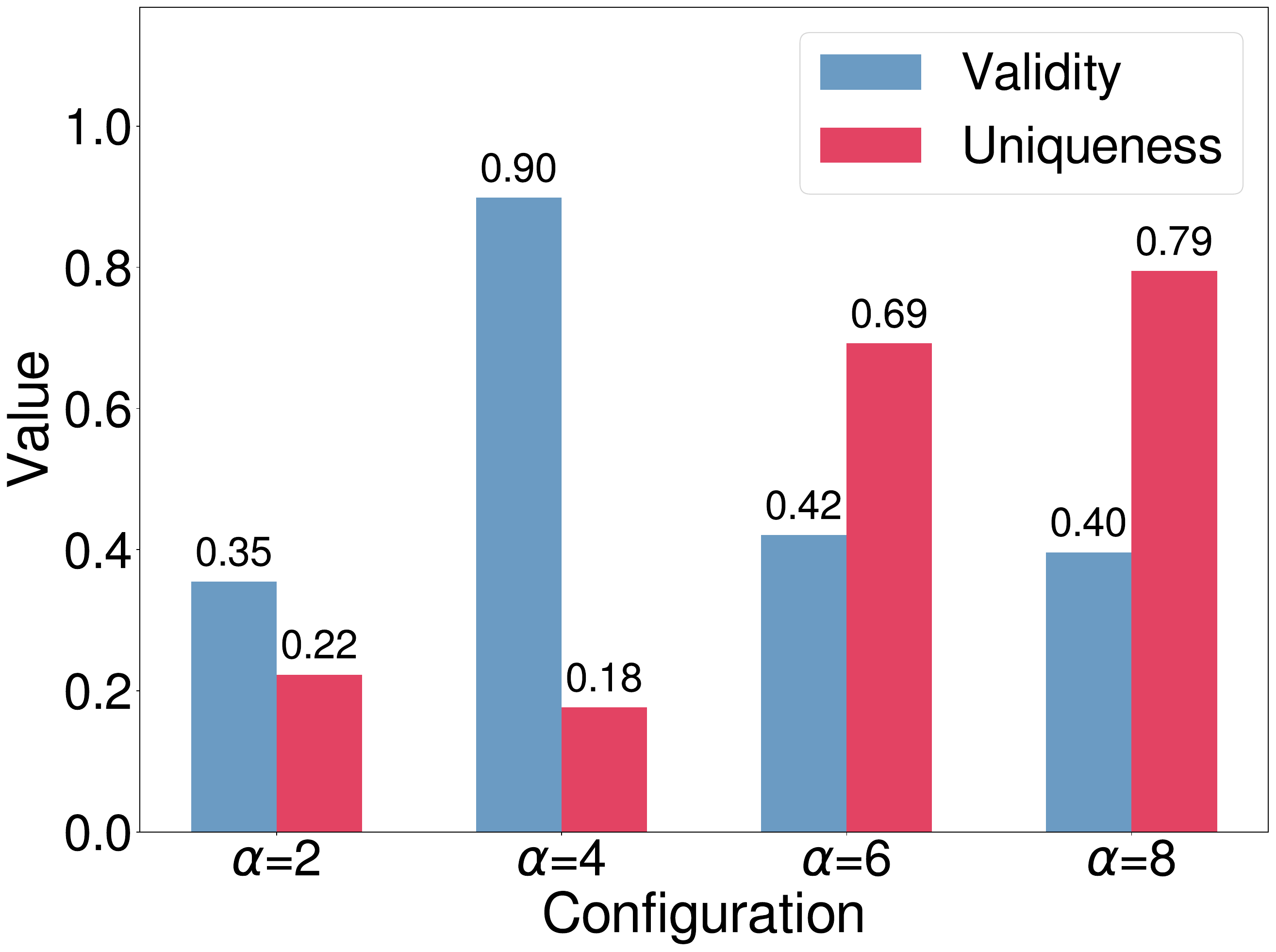}
    \caption{Validity-Uniqueness Trade-off at different values of $\alpha$. Note that validity represents the quality of generated samples, while uniqueness represents the diversity of generated samples.}
    \label{fig: validity-uniqueness-trade-off}
\end{figure}

\paragraph{Mitigation of Sample Bias}

Figure \ref{fig: tanimoto-plot} shows that the distance distribution of the generated compounds to the anchor compounds is similar to that of the positive compounds. We calculated the distance over the fingerprint domain to consider the patterns of molecular substructure. The result implies that the generated and positive compounds have different substructures from the anchor compounds to the same extent. However, Figure \ref{fig: tanimoto-plot} compares \textit{``the relative distances''} of the generated and positive compounds to the anchor compounds \textit{``at the substructure level,''} making it difficult to directly compare the absolute distances to each other at the holistic level.

Figure \ref{fig: kde-plot} visualizes the overlapping representations among the anchor, positive, and generated compounds, \textit{``directly comparing their absolute distances at the holistic level.''} To do this, we extracted SMILES word embeddings (\textit{e.g.,} C, N, F, =, +, [, ], etc.) using Word2Vec \cite{mikolov2013efficient} and defined the holistic representation of each molecule as the summation of these word embeddings. We then projected the holistic representation of each molecule into a 2-dimensional space by t-SNE \cite{van2008visualizing}. Since t-SNE embeddings preserve pairwise similarities of high-dimensional data as neighboring points in a low-dimension, it allows for direct comparison of absolute distances between samples. Finally, we applied Gaussian kernel density estimation (KDE) to visualize the distribution of t-SNE embeddings.

The results from Figures \ref{fig: tanimoto-plot} and \ref{fig: kde-plot} indicate that the generated compounds are more similar to the positive compounds than to the anchor compounds, both relatively and absolutely, and at substructural and holistic levels. This suggests that Repurformer successfully addressed the sample bias problem, creating substitutes for anchor compounds that resemble positive compounds.



\begin{figure}[t]
    \centering
    \includegraphics[width=0.95\columnwidth]{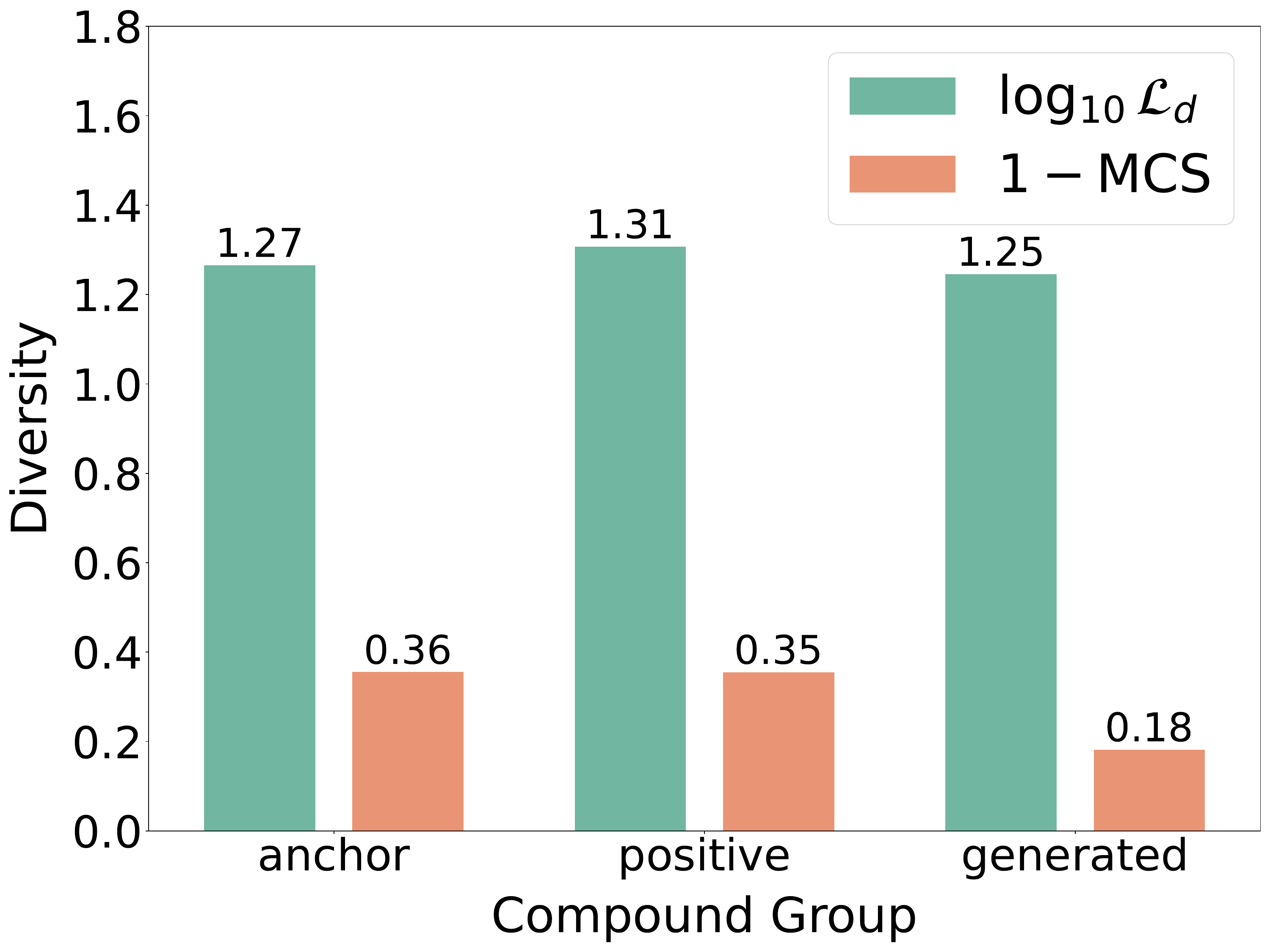}
    \caption{Internal Diversity per Compound Group. $\text{log}_{10}\mathcal{L}_{d}$ measures the syntactic difference of SMILES strings, while MCS distance (1-MCS) measures the semantic dissimilarity at the atomic level.}
    \label{fig: internal-diversity-combine}
\end{figure}

\paragraph{Existence of Mode Collapse}
Mode collapse refers to a phenomenon where the generative model creates high-quality samples at the expense of in-distribution diversity \cite{adiga2018tradeoff}. In this section, we demonstrate that Repurformer suffers from mode collapse and thus the \textit{``internal''} diversity of generated compounds is relatively lower than anchor and positive compounds. 

Figure \ref{fig: validity-uniqueness-trade-off} illustrates the negative relationship between the validity and uniqueness of the generated compounds by different values of $\alpha$. Validity represents the ratio of samples that can be depicted as 2D molecular drawings by RDKit (\textit{i.e.,} the quality of generation), while uniqueness represents the ratio of non-duplicated samples (\textit{i.e.,} the diversity of generation). As $\alpha$ increases, we observe that uniqueness increases but validity decreases. This is an expected outcome given that the low-frequency signals represent global structure whereas the high-frequency signals represent local structure. For example, low-frequency signals (\textit{i.e.,} lower $\alpha$) focus on the most fundamental structures, increasing the validity of generated compounds but reducing their uniqueness. Conversely, high-frequency signals (\textit{i.e.,} higher $\alpha$) focus on local details, increasing the uniqueness of generated compounds but reducing their structural validity. In short, Figure \ref{fig: validity-uniqueness-trade-off} demonstrates that Repurformer may be susceptible to the validity-uniqueness trade-off, \textit{i.e.,} mode collapse, and thus $\alpha$ must be carefully selected.

Figure \ref{fig: internal-diversity-combine} describes the \textit{``internal''} diversity of valid compounds generated by Repurformer with $\alpha=4$, along with anchor and positive compounds that share the same target proteins with the generated ones. Following \citet{pereira2021diversity}, we evaluated the internal diversity within each compound group using two metrics: Levenshtein distance ($\mathcal{L}_{d}$) and maximum common substructure (MCS). The Levenshtein distance \cite{levenshtein1966binary}, also known as edit distance, measures the difference between two SMILES strings by calculating the minimum number of insertions, deletions, and replacements needed to make the strings identical. On the other hand, the MCS \cite{cao2008maximum} measures the ratio of the number of atoms in the maximum common substructure of two compounds to their total number of atoms. Since the MCS represents a similarity score normalized between 0 and 1, the MCS distance can be obtained by 1-MCS, which captures the atom-level dissimilarities of compounds. The diversity of each compound group was computed by averaging the pairwise distances of all compounds. The result indicates that the internal diversity of the generated compounds is lower than that of the anchor and positive compounds, suggesting mode collapse in Repurformer. 

Note that the existence of mode collapse does not contradict the mitigation of sample bias. Mode collapse refers to less internal diversity among generated compounds, while mitigating sample bias involves creating substitutes for anchor compounds that resemble positive ones, thus increasing diversity between the anchor and generated compounds.

\section{Limitations} \label{sec: limitation}
This study has some limitations. First, due to inconsistencies between the tokens in our dataset and those we borrowed from previous research, some generated outputs contained <UNK> tokens, which had to be excluded. Second, the study lacks experiments on binding affinity, which are necessary to evaluate how strongly the generated compounds bind to proteins. These limitations must be addressed in future research.

\section{Concluding Remarks} \label{sec: conclusion}
In this study, we introduced Repurformer, a novel model designed to address the sample bias problem in \textit{de novo} molecule generation by leveraging multi-hop relationships. Repurformer integrates bi-directional pretraining with Fast Fourier Transform and low-pass filtering, to capture complex interactions between proteins and compounds. This approach focuses on low-frequency components, corresponding to longer propagation through multi-hop protein-compound interactions. The results show that Repurformer successfully generates valid and diverse molecules.

Building on these positive results, there are several promising directions for future improvement. Enhancing the backbone architecture by incorporating advanced models like diffusion or graph neural networks and techniques such as contrastive learning could further improve Repurformer's ability to capture multi-hop protein-compound interactions. The results from Figures \ref{fig: validity-uniqueness-trade-off} and \ref{fig: internal-diversity-combine} also suggest promising directions to improve Repurformer such as leveraging reinforcement learning to maximize diversity rewards or introducing Wasserstein loss to address mode collapse. Additionally, while our current experiments have shown the potential of Repurformer, it is critical to validate its applicability in real-world scenarios. Therefore, we need to verify the performance of Repurformer on existing drug repurposing cases. Considering these aspects will strengthen the practical implications and utilities of Repurformer.




\bibliography{custom}

\begin{thebibliography}{42}
\providecommand{\natexlab}[1]{#1}

\bibitem[{Adiga et~al.(2018)Adiga, Attia, Chang, and Tandon}]{adiga2018tradeoff}
Sudarshan Adiga, Mohamed~Adel Attia, Wei-Ting Chang, and Ravi Tandon. 2018.
\newblock On the tradeoff between mode collapse and sample quality in generative adversarial networks.
\newblock In \emph{2018 IEEE global conference on signal and information processing (GlobalSIP)}, pages 1184--1188. IEEE.

\bibitem[{Bagal et~al.(2022)Bagal, Aggarwal, Vinod, and Priyakumar}]{bagal2022}
Viraj Bagal, Rishal Aggarwal, P.~K. Vinod, and U.~Deva Priyakumar. 2022.
\newblock \href {https://doi.org/10.1021/acs.jcim.1c00600} {{MolGPT}: {Molecular} generation using a transformer-decoder model}.
\newblock \emph{Journal of Chemical Information and Modeling}, 62(9):2064--2076.

\bibitem[{Bickerton et~al.(2012)Bickerton, Paolini, Besnard, Muresan, and Hopkins}]{bickerton2012}
G.~Richard Bickerton, Gaia~V. Paolini, J{\'e}r{\'e}my Besnard, Sorel Muresan, and Andrew~L. Hopkins. 2012.
\newblock \href {https://doi.org/10.1038/nchem.1243} {Quantifying the chemical beauty of drugs}.
\newblock \emph{Nature Chemistry}, 4(2):90--98.

\bibitem[{Brigham(1988)}]{brigham1988fast}
E~Oran Brigham. 1988.
\newblock \emph{The fast Fourier transform and its applications}.
\newblock Prentice-Hall, Inc.

\bibitem[{Cao et~al.(2008)Cao, Jiang, and Girke}]{cao2008maximum}
Yiqun Cao, Tao Jiang, and Thomas Girke. 2008.
\newblock A maximum common substructure-based algorithm for searching and predicting drug-like compounds.
\newblock \emph{Bioinformatics}, 24(13):i366--i374.

\bibitem[{Cooley and Tukey(1965)}]{cooley1965algorithm}
James~W Cooley and John~W Tukey. 1965.
\newblock An algorithm for the machine calculation of complex fourier series.
\newblock \emph{Mathematics of computation}, 19(90):297--301.

\bibitem[{Costen et~al.(1996)Costen, Parker, and Craw}]{costen1996effects}
Nicholas~P Costen, Denis~M Parker, and Ian Craw. 1996.
\newblock Effects of high-pass and low-pass spatial filtering on face identification.
\newblock \emph{Perception \& psychophysics}, 58:602--612.

\bibitem[{Dai et~al.(2018)Dai, Tian, Dai, Skiena, and Song}]{dai2018}
Hanjun Dai, Yingtao Tian, Bo~Dai, Steven Skiena, and Le~Song. 2018.
\newblock Syntax-directed variational autoencoder for molecule generation.
\newblock In \emph{31st {Conference} on {Neural} {Information} {Processing} {Systems} ({NIPS} 2017)}.

\bibitem[{Ertl et~al.(2008)Ertl, Roggo, and Schuffenhauer}]{ertl2008}
Peter Ertl, Silvio Roggo, and Ansgar Schuffenhauer. 2008.
\newblock \href {https://doi.org/10.1021/ci700286x} {Natural {{Product-likeness Score}} and {{Its Application}} for {{Prioritization}} of {{Compound Libraries}}}.
\newblock \emph{Journal of Chemical Information and Modeling}, 48(1):68--74.

\bibitem[{Ertl and Schuffenhauer(2009)}]{ertl2009}
Peter Ertl and Ansgar Schuffenhauer. 2009.
\newblock \href {https://doi.org/10.1186/1758-2946-1-8} {Estimation of synthetic accessibility score of drug-like molecules based on molecular complexity and fragment contributions}.
\newblock \emph{Journal of Cheminformatics}, 1(1):8.

\bibitem[{Gilson et~al.(2016)Gilson, Liu, Baitaluk, Nicola, Hwang, and Chong}]{gilson2016}
Michael~K. Gilson, Tiqing Liu, Michael Baitaluk, George Nicola, Linda Hwang, and Jenny Chong. 2016.
\newblock \href {https://doi.org/10.1093/nar/gkv1072} {{BindingDB} in 2015: {A} public database for medicinal chemistry, computational chemistry and systems pharmacology}.
\newblock \emph{Nucleic Acids Research}, 44(D1):D1045--D1053.

\bibitem[{Goodfellow et~al.(2014)Goodfellow, Pouget-Abadie, Mirza, Xu, Warde-Farley, Ozair, Courville, and Bengio}]{GoodfellowGAN2014}
Ian Goodfellow, Jean Pouget-Abadie, Mehdi Mirza, Bing Xu, David Warde-Farley, Sherjil Ozair, Aaron Courville, and Yoshua Bengio. 2014.
\newblock \href {https://proceedings.neurips.cc/paper_files/paper/2014/file/5ca3e9b122f61f8f06494c97b1afccf3-Paper.pdf} {Generative adversarial nets}.
\newblock In \emph{Advances in Neural Information Processing Systems}, volume~27. Curran Associates, Inc.

\bibitem[{Grechishnikova(2021)}]{grechishnikovaTransformerNeuralNetwork2021}
Daria Grechishnikova. 2021.
\newblock \href {https://doi.org/10.1038/s41598-020-79682-4} {Transformer neural network for protein-specific de novo drug generation as a machine translation problem}.
\newblock \emph{Scientific Reports}, 11(1):321.

\bibitem[{Guimaraes et~al.(2018)Guimaraes, Sanchez-Lengeling, Outeiral, Farias, and Aspuru-Guzik}]{guimaraes2018}
Gabriel~Lima Guimaraes, Benjamin Sanchez-Lengeling, Carlos Outeiral, Pedro Luis~Cunha Farias, and Alán Aspuru-Guzik. 2018.
\newblock \href {http://arxiv.org/abs/1705.10843} {Objective-reinforced generative adversarial networks ({ORGAN}) for sequence generation models}.
\newblock \emph{arXiv preprint}.
\newblock ArXiv:1705.10843 [cs, stat].

\bibitem[{Gómez-Bombarelli et~al.(2018)Gómez-Bombarelli, Wei, Duvenaud, Hernández-Lobato, Sánchez-Lengeling, Sheberla, Aguilera-Iparraguirre, Hirzel, Adams, and Aspuru-Guzik}]{gomez-bombarelli2018}
Rafael Gómez-Bombarelli, Jennifer~N. Wei, David Duvenaud, José~Miguel Hernández-Lobato, Benjamín Sánchez-Lengeling, Dennis Sheberla, Jorge Aguilera-Iparraguirre, Timothy~D. Hirzel, Ryan~P. Adams, and Alán Aspuru-Guzik. 2018.
\newblock \href {https://doi.org/10.1021/acscentsci.7b00572} {Automatic chemical design using a data-driven continuous representation of molecules}.
\newblock \emph{ACS Central Science}, 4(2):268--276.

\bibitem[{Heckbert(1995)}]{heckbert1995fourier}
Paul Heckbert. 1995.
\newblock Fourier transforms and the fast fourier transform (fft) algorithm.
\newblock \emph{Computer Graphics}, 2(1995):15--463.

\bibitem[{Ho et~al.(2020)Ho, Jain, and Abbeel}]{Ho2020denoising}
Jonathan Ho, Ajay Jain, and Pieter Abbeel. 2020.
\newblock \href {https://arxiv.org/abs/2006.11239} {Denoising diffusion probabilistic models}.
\newblock \emph{Preprint}, arXiv:2006.11239.

\bibitem[{Honda et~al.(2019)Honda, Shi, and Ueda}]{honda2019smiles}
Shion Honda, Shoi Shi, and Hiroki~R Ueda. 2019.
\newblock Smiles transformer: Pre-trained molecular fingerprint for low data drug discovery.
\newblock \emph{arXiv preprint arXiv:1911.04738}.

\bibitem[{Huang et~al.(2023)Huang, Zhang, Xu, and Wong}]{huang2023}
Lei Huang, Hengtong Zhang, Tingyang Xu, and Ka-Chun Wong. 2023.
\newblock \href {https://doi.org/10.1609/aaai.v37i4.25639} {{MDM}: {Molecular} diffusion model for {3D} molecule generation}.
\newblock \emph{Proceedings of the AAAI Conference on Artificial Intelligence}, 37(4):5105--5112.

\bibitem[{Jin et~al.(2018)Jin, Barzilay, and Jaakkola}]{jin2018}
Wengong Jin, Regina Barzilay, and Tommi Jaakkola. 2018.
\newblock \href {https://proceedings.mlr.press/v80/jin18a.html} {Junction tree variational autoencoder for molecular graph generation}.
\newblock In \emph{Proceedings of the 35th {International} {Conference} on {Machine} {Learning}}, pages 2323--2332. PMLR.
\newblock ISSN: 2640-3498.

\bibitem[{Kingma and Welling(2022)}]{kingmaAutoencodingVariationalBayes2022}
Diederik~P. Kingma and Max Welling. 2022.
\newblock \href {https://arxiv.org/abs/1312.6114} {Auto-encoding variational bayes}.
\newblock \emph{Preprint}, arxiv:1312.6114.

\bibitem[{Lee-Thorp et~al.(2021)Lee-Thorp, Ainslie, Eckstein, and Ontanon}]{lee2021fnet}
James Lee-Thorp, Joshua Ainslie, Ilya Eckstein, and Santiago Ontanon. 2021.
\newblock Fnet: Mixing tokens with fourier transforms.
\newblock \emph{arXiv preprint arXiv:2105.03824}.

\bibitem[{Levenshtein et~al.(1966)}]{levenshtein1966binary}
Vladimir~I Levenshtein et~al. 1966.
\newblock Binary codes capable of correcting deletions, insertions, and reversals.
\newblock In \emph{Soviet physics doklady}, volume~10, pages 707--710. Soviet Union.

\bibitem[{Lin(2004)}]{lin2004b}
Chin-Yew Lin. 2004.
\newblock {{ROUGE}}: {{A}} package for automatic evaluation of summaries.
\newblock In \emph{Text {{Summarization Branches Out}}}, pages 74--81, Barcelona, Spain. Association for Computational Linguistics.

\bibitem[{Lipinski et~al.(2012)Lipinski, Lombardo, Dominy, and Feeney}]{lipinski2012experimental}
Christopher~A Lipinski, Franco Lombardo, Beryl~W Dominy, and Paul~J Feeney. 2012.
\newblock Experimental and computational approaches to estimate solubility and permeability in drug discovery and development settings.
\newblock \emph{Advanced drug delivery reviews}, 64:4--17.

\bibitem[{Maziarka et~al.(2020)Maziarka, Pocha, Kaczmarczyk, Rataj, Danel, and Warchoł}]{maziarka2020}
Łukasz Maziarka, Agnieszka Pocha, Jan Kaczmarczyk, Krzysztof Rataj, Tomasz Danel, and Michał Warchoł. 2020.
\newblock \href {https://doi.org/10.1186/s13321-019-0404-1} {Mol-{CycleGAN}: a generative model for molecular optimization}.
\newblock \emph{Journal of Cheminformatics}, 12(1):2.

\bibitem[{Mikolov et~al.(2013)Mikolov, Chen, Corrado, and Dean}]{mikolov2013efficient}
Tomas Mikolov, Kai Chen, Greg Corrado, and Jeffrey Dean. 2013.
\newblock Efficient estimation of word representations in vector space.
\newblock \emph{arXiv preprint arXiv:1301.3781}.

\bibitem[{Munson et~al.(2024)Munson, Chen, Bogosian, Kreisberg, Licon, Abagyan, Kuenzi, and Ideker}]{munson2024}
Brenton~P. Munson, Michael Chen, Audrey Bogosian, Jason~F. Kreisberg, Katherine Licon, Ruben Abagyan, Brent~M. Kuenzi, and Trey Ideker. 2024.
\newblock \href {https://doi.org/10.1038/s41467-024-47120-y} {De novo generation of multi-target compounds using deep generative chemistry}.
\newblock \emph{Nature Communications}, 15(1):3636.

\bibitem[{Papineni et~al.(2002)Papineni, Roukos, Ward, and Zhu}]{papineni2002}
Kishore Papineni, Salim Roukos, Todd Ward, and Wei-Jing Zhu. 2002.
\newblock \href {https://doi.org/10.3115/1073083.1073135} {Bleu: {{A}} method for automatic evaluation of machine translation}.
\newblock In \emph{Proceedings of the 40th {{Annual Meeting}} of the {{Association}} for {{Computational Linguistics}}}, pages 311--318, Philadelphia, Pennsylvania, USA. Association for Computational Linguistics.

\bibitem[{Pereira et~al.(2021)Pereira, Abbasi, Ribeiro, and Arrais}]{pereira2021diversity}
Tiago Pereira, Maryam Abbasi, Bernardete Ribeiro, and Joel~P Arrais. 2021.
\newblock Diversity oriented deep reinforcement learning for targeted molecule generation.
\newblock \emph{Journal of cheminformatics}, 13(1):21.

\bibitem[{Pollack(1948)}]{pollack1948effects}
Irwin Pollack. 1948.
\newblock Effects of high pass and low pass filtering on the intelligibility of speech in noise.
\newblock \emph{The Journal of the Acoustical Society of America}, 20(3):259--266.

\bibitem[{Qian et~al.(2022)Qian, Lin, Zhao, Tu, and Xu}]{qianAlphaDrugProteinTarget2022}
Hao Qian, Cheng Lin, Dengwei Zhao, Shikui Tu, and Lei Xu. 2022.
\newblock \href {https://doi.org/10.1093/pnasnexus/pgac227} {{{AlphaDrug}}: Protein target specific de novo molecular generation}.
\newblock \emph{PNAS Nexus}, 1(4):pgac227.

\bibitem[{Rao and Yip(2014)}]{rao2014discrete}
K~Ramamohan Rao and Ping Yip. 2014.
\newblock \emph{Discrete cosine transform: algorithms, advantages, applications}.
\newblock Academic press.

\bibitem[{Rao et~al.(2019)Rao, Bhattacharya, Thomas, Duan, Chen, Canny, Abbeel, and Song}]{rao2019evaluating}
Roshan Rao, Nicholas Bhattacharya, Neil Thomas, Yan Duan, Peter Chen, John Canny, Pieter Abbeel, and Yun Song. 2019.
\newblock Evaluating protein transfer learning with tape.
\newblock \emph{Advances in neural information processing systems}, 32.

\bibitem[{Sanchez-Lengeling et~al.(2017)Sanchez-Lengeling, Outeiral, and Guimaraes}]{sanchez-lengeling2017}
Benjamin Sanchez-Lengeling, Carlos Outeiral, and Gabriel~L Guimaraes. 2017.
\newblock \href {https://doi.org/10.26434/chemrxiv.5309668.v3} {Optimizing distributions over molecular space. {An} objective-reinforced generative adversarial network for inverse-design chemistry ({ORGANIC})}.

\bibitem[{Tamkin et~al.(2020)Tamkin, Jurafsky, and Goodman}]{tamkin2020language}
Alex Tamkin, Dan Jurafsky, and Noah Goodman. 2020.
\newblock Language through a prism: A spectral approach for multiscale language representations.
\newblock \emph{Advances in Neural Information Processing Systems}, 33:5492--5504.

\bibitem[{Tan et~al.(2022)Tan, Gao, and Li}]{tan2022}
Cheng Tan, Zhangyang Gao, and Stan~Z. Li. 2022.
\newblock \href {http://arxiv.org/abs/2202.04829} {Target-aware molecular graph generation}.
\newblock \emph{arXiv preprint}.
\newblock ArXiv:2202.04829 [cs].

\bibitem[{Van~der Maaten and Hinton(2008)}]{van2008visualizing}
Laurens Van~der Maaten and Geoffrey Hinton. 2008.
\newblock Visualizing data using t-sne.
\newblock \emph{Journal of machine learning research}, 9(11).

\bibitem[{Vaswani et~al.(2017)Vaswani, Shazeer, Parmar, Uszkoreit, Jones, Gomez, Kaiser, and Polosukhin}]{Vaswani2017attention}
Ashish Vaswani, Noam Shazeer, Niki Parmar, Jakob Uszkoreit, Llion Jones, Aidan~N. Gomez, Lukasz Kaiser, and Illia Polosukhin. 2017.
\newblock \href {https://arxiv.org/abs/1706.03762} {Attention is all you need}.
\newblock \emph{Preprint}, arXiv:1706.03762.

\bibitem[{Wildman and Crippen(1999)}]{wildman1999}
Scott~A. Wildman and Gordon~M. Crippen. 1999.
\newblock \href {https://doi.org/10.1021/ci990307l} {Prediction of physicochemical parameters by atomic contributions}.
\newblock \emph{Journal of Chemical Information and Computer Sciences}, 39(5):868--873.

\bibitem[{Wu et~al.(2016)Wu, Schuster, Chen, Le, Norouzi, Macherey, Krikun, Cao, Gao, Macherey, Klingner, Shah, Johnson, Liu, Kaiser, Gouws, Kato, Kudo, Kazawa, Stevens, Kurian, Patil, Wang, Young, Smith, Riesa, Rudnick, Vinyals, Corrado, Hughes, and Dean}]{wuGoogleNeuralMachine2016}
Yonghui Wu, Mike Schuster, Zhifeng Chen, Quoc~V. Le, Mohammad Norouzi, Wolfgang Macherey, Maxim Krikun, Yuan Cao, Qin Gao, Klaus Macherey, Jeff Klingner, Apurva Shah, Melvin Johnson, Xiaobing Liu, {\L}ukasz Kaiser, Stephan Gouws, Yoshikiyo Kato, Taku Kudo, Hideto Kazawa, Keith Stevens, George Kurian, Nishant Patil, Wei Wang, Cliff Young, Jason Smith, Jason Riesa, Alex Rudnick, Oriol Vinyals, Greg Corrado, Macduff Hughes, and Jeffrey Dean. 2016.
\newblock \href {https://arxiv.org/abs/1609.08144} {Google's neural machine translation system: {{Bridging}} the gap between human and machine translation}.
\newblock \emph{Preprint}, arxiv:1609.08144.

\bibitem[{Xu et~al.(2023)Xu, Powers, Dror, Ermon, and Leskovec}]{xu2023}
Minkai Xu, Alexander~S. Powers, Ron~O. Dror, Stefano Ermon, and Jure Leskovec. 2023.
\newblock \href {https://proceedings.mlr.press/v202/xu23n.html} {Geometric latent diffusion models for {3D} molecule generation}.
\newblock In \emph{Proceedings of the 40th {International} {Conference} on {Machine} {Learning}}, pages 38592--38610. PMLR.
\newblock ISSN: 2640-3498.

\end{thebibliography}

\appendix
\onecolumn


\section{Supplementary Materials}

\begin{figure}[ht]
    \centering
    \includegraphics[width=\linewidth]{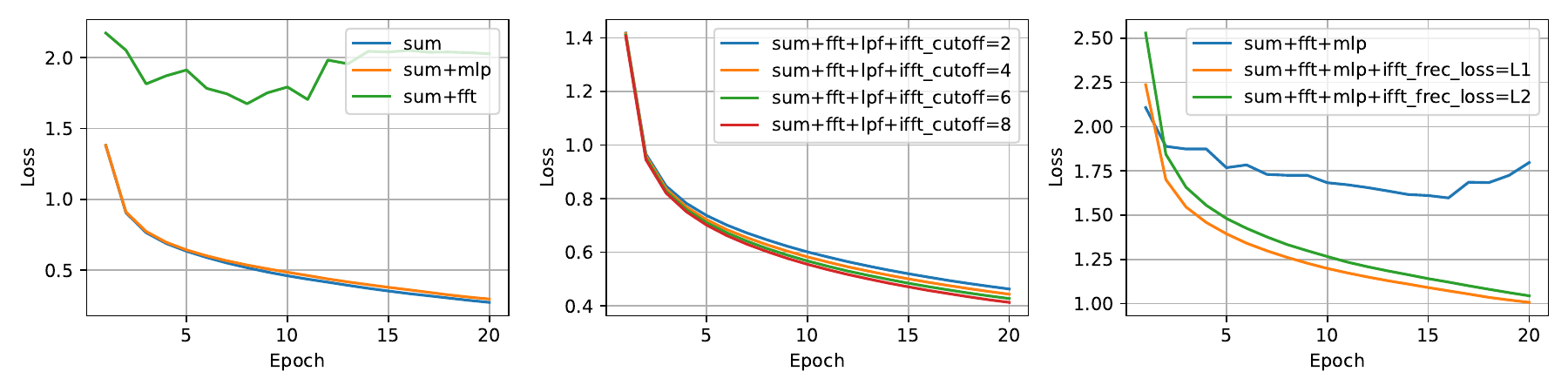}
    \caption{Comparison of Training Performance with Different Configurations. When embedding vectors from protein and compound encoders are mapped to the frequency domain using Fourier Transform (FFT), training performance does not improve unless they are transformed back to the original domain with an inverse Fourier Transform (iFFT). This indicates that applying FFT in the latent space leads to alignment issues between the encoders and the compound decoder.}
    \label{fig: loss-per-configuration}
\end{figure}

\begin{figure}[ht]
    \centering
    \includegraphics[width=0.45\linewidth]{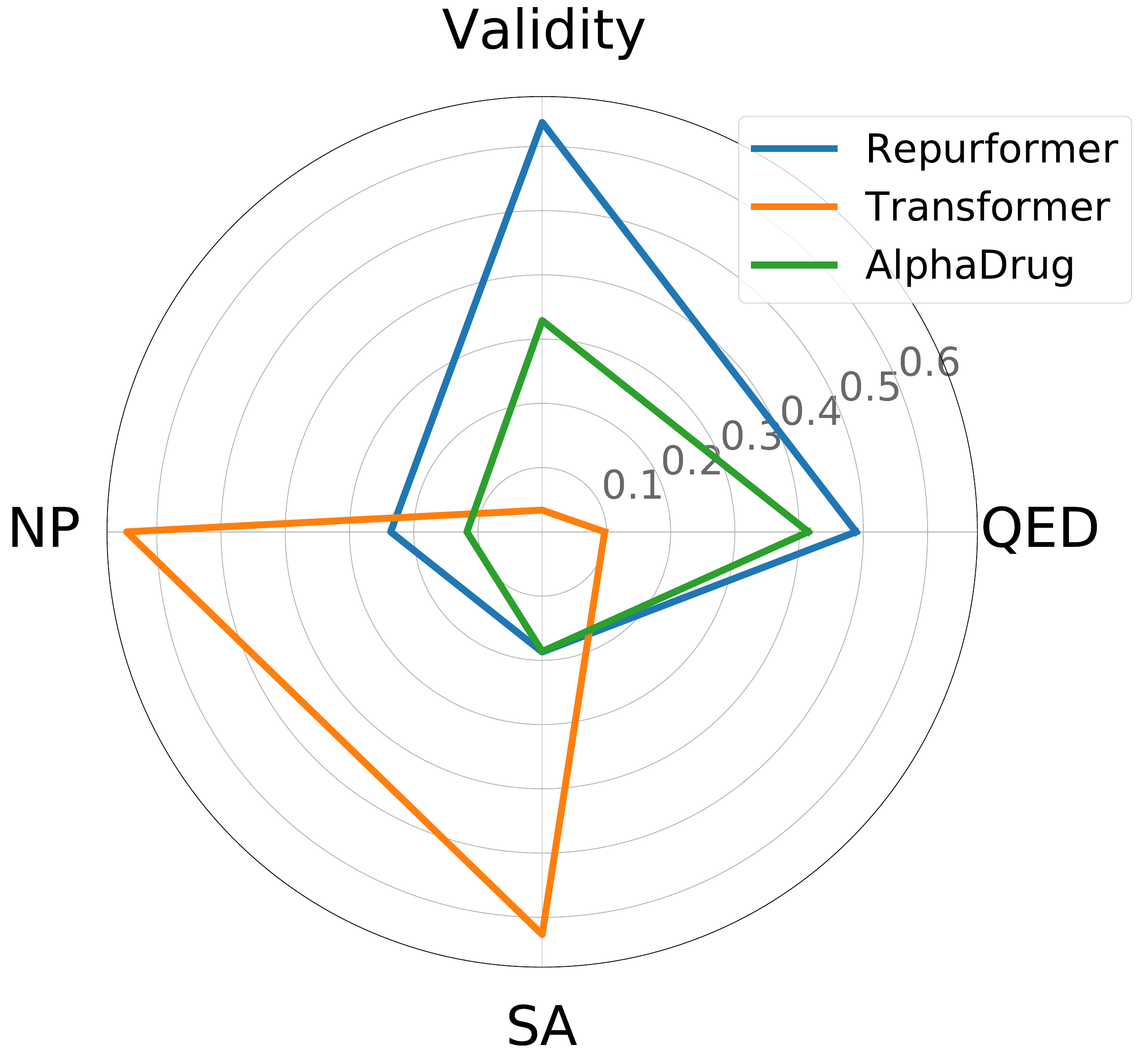}
    \caption{Comparison of Validity and Drug-Likeness Metrics. Validity, QED, SA, and NP scores were normalized to the same scale. Although the Transformer-based model \cite{grechishnikovaTransformerNeuralNetwork2021} showed relatively higher SA and NP scores, its validity is extremely low. This indicates that the compounds generated by the Transformer-based model are not of sufficient quality to be considered as drug candidates.}
    \label{fig: radar-chart}
\end{figure}

\end{document}